\documentclass[11 pt]{article}
\usepackage[margin=1in]{geometry}
\usepackage{hyperref}       
\usepackage{url}            
\usepackage{booktabs}       
\usepackage{amsfonts}       
\usepackage{nicefrac}       
\usepackage{microtype}      
\usepackage{amstext}
\usepackage{amsthm}
\usepackage{amssymb}
\usepackage{amsmath}
\usepackage{makecell}
\usepackage[numbers]{natbib}
\usepackage{caption}
\usepackage{subcaption}
\usepackage{graphicx}
\usepackage[export]{adjustbox}

\usepackage{multirow}

\usepackage{algorithm}
\usepackage{algorithmic}
\usepackage{comment}

\usepackage{titlesec} 
\titleformat{\subsubsection}[runin]{\normalfont\bfseries}{\thesubsubsection.}{3pt}{}

\usepackage{enumitem}

\usepackage{xcolor}
\newcommand{\yc}[1]{{\textcolor{blue}{\bf [Yudong: #1]}}}
\newcommand{\yz}[1]{{\textcolor{teal}{\bf [Yuqian: #1]}}}
\newcommand{\ld}[1]{{\textcolor{red}{\bf [Lijun: #1]}}}
\newcommand{\newcontent}[1]{\textcolor{black}{#1}}
\newcommand{\newcontentyc}[1]{\textcolor{black}{#1}}
\newcommand{\newcontentyz}[1]{{\textcolor{black}{#1}}}
\newcommand{\td}[1]{{\textcolor{magenta}{\bf }}}

\renewcommand{\yc}[1]{{\textcolor{blue}{\bf}}}
\renewcommand{\yz}[1]{{\textcolor{teal}{\bf}}}
\renewcommand{\ld}[1]{{\textcolor{red}{\bf}}}

\newtheorem{thm}{\protect\theoremname}
\newtheorem{defn}{\protect\definitionname}
\newtheorem{lem}{\protect\lemmaname}
\newtheorem*{lem*}{\protect\lemmaname}

\newtheorem{cor}{Corollary}

\theoremstyle{definition}
\newtheorem{exmp}{Example}
\makeatother

\providecommand{\definitionname}{Definition}
\providecommand{\lemmaname}{Lemma}
\providecommand{\remarkname}{Remark}
\providecommand{\theoremname}{Theorem}

\global\long\def\fronorm#1{\|#1\|_{\tiny \mbox{F}}}

\global \long \def\norm #1{\|#1\|}
\global\long\def\twonorm#1{\|#1\|_{2}}

\global\long\def\opnorm#1{\|#1\|_{\tiny \mbox{op}}}

\global\long\def\nucnorm#1{\|#1\|_{\tiny \mbox{nuc}}}

\global\long\def\infnorm#1{\|#1\|_{\infty}}

\global\long\def\abs#1{\left|#1\right|}

\global\long\def\real{\mathbb{R}}

\global\long\def\integer{\mathbb{Z}}

\global\long\def\inprod#1#2{\langle#1,#2\rangle}

\global\long\def\dm{d}

\global\long\def\Amap{\mathcal{A}}

\global\long\def\rank{\mbox{rank}}

\global\long\def\proj{\mathcal{P}}

\global \long \def \Prob {\mathbb{P}}

\global \long \def \Exs {\mathbb{E}}

\global \long \def \Amap {\mathcal{A}}
\global \long \def \logPart {\psi}

\global\long \def \loss{\mathcal{L}}
\global \long \def \nsample{n}
\global \long \def \d {d}
\global\long \def \truX {X^\natural}
\global \long \def \trurank {r^\natural}
\global \long \def \rsm{\beta}
\global \long \def \rsc {\alpha}
\global \long \def \nablasmall {\epsilon_{\nabla}}

\global \long \def \bigO{\mathcal{O}} 
\global \long \def \conset{\mathcal{C}}
\global \long \def \Asense{A}
\DeclareMathOperator*{\argmin}{arg\,min}
\newcommand{\frotruX}{\xi_0}

\newcommand{\rscerror}{\epsilon_{\rsc}}
\newcommand{\rsmerror}{\epsilon_{\rsm}}
\newcommand{\condn}{\kappa}

\title{Low-rank matrix recovery with non-quadratic loss: projected gradient method and  regularity projection oracle}

\author{Lijun Ding\footnote{Department of Mathematics, University of Washington, Seattle WA 98195, USA; \texttt{ljding@uw.edu}},  
	Yuqian Zhang\footnote{Department of Electrical and Computer Engineering, Rutgers University, Piscataway, NJ 08854, USA;
		\texttt{yqz.zhang@rutgers.edu}}, 
	and Yudong Chen\footnote{Department of Computer Sciences,
University of Wisconsin-Madison, Madison, WI 53706, USA; \texttt{yudong.chen@wisc.edu}}
}
\begin{document}

\maketitle

\begin{abstract}
Existing results for low-rank matrix recovery largely focus on quadratic loss, which enjoys favorable properties such as restricted strong convexity/smoothness (RSC/RSM) and well conditioning over all low rank matrices. However, many interesting problems involve more general, non-quadratic losses, which do not satisfy such properties.
For these problems, standard nonconvex approaches such as rank-constrained projected gradient descent (a.k.a.\ iterative hard thresholding) and  Burer-Monteiro factorization could have poor empirical performance, and there is no satisfactory theory guaranteeing global and fast convergence for these algorithms. 
		
In this paper, we show that a critical component in provable low-rank recovery with non-quadratic loss is a regularity projection oracle. This oracle restricts iterates to low-rank matrices within an appropriate bounded set, over which the loss function is well behaved and satisfies a set of approximate RSC/RSM conditions. Accordingly, we analyze an (averaged) projected gradient method equipped with such an oracle, and prove that it converges globally and linearly. Our results apply to a wide range of non-quadratic low-rank estimation problems including one bit matrix sensing/completion,  individualized rank aggregation, and more broadly generalized linear models with rank constraints.
\end{abstract}

	\section{Introduction} 
	
	In this paper, we consider the problem of \emph{rank-constrained generalized linear model} (RGLM), where the goal is to recover a 
	rank-$\trurank$ ground truth matrix $\truX\in \real^{\dm_1\times \dm_2}$ from
	independent data $(y_i,A_i)\in \real \times \real^{\dm_1\times \dm_2}, i=1,\dots ,\nsample$ generated as follows.
	Given the measurement matrix $\Asense_i$, the response $y_i$ follows a generalized linear model \cite{friedman2010regularization} with the exponential family distribution
	\begin{equation}\label{eq: glm}
	\begin{aligned} 
	\Prob(y_i \mid \Asense_i) \propto \exp \left\{ \frac{y_i \inprod{\Asense_i}{\truX} -\logPart\left( \inprod{\Asense_i}{\truX}\right)}{c(\sigma)}  \right\},
	\end{aligned} 
	\end{equation}
	where $\psi$ is some convex log partition function that is twice continuously differentiable and $c(\sigma)$ is a function measuring the noise level. 
	Examples of RGLM include matrix sensing with Gaussian noise \cite{candes2011tight}, one-bit matrix sensing (a generalization of one-bit compressed sensing \cite{boufounos20081}), noisy matrix completion \cite{candes2010matrix}, 
	one-bit matrix completion \cite{davenport20141}, 
	individualized rank aggregation from pairwise comparison \cite{lu2015individualized}, and so on. 
	\paragraph{Nonconvex formulation and regularity projection oracle} 
	
	The above problem can be cast as a rank-constrained nonconvex optimization problem as follows:
	\begin{equation}\label{opt: mainProblem}
	\begin{array}{ll}
	\mbox{minimize}_{X\in \real ^{\d_1\times \d_2}} & \loss (X):\,= \frac{1}{n} \sum_{i=1}^\nsample \big[ \logPart\left( \inprod{X}{\Asense_i}\right) -y_i\inprod{\Asense_i}{X} \big] \\
	\mbox{subject to} &  \rank(X)\leq \trurank,\, X \in \conset. 
	\end{array}
	\end{equation}
	Here, $ \loss$ is the negative log-likelihood function, which is convex due to the convexity of $\psi$. Whenever the distribution of $ \{y_i\}  $ is non-Gaussian, $ \loss $ is \emph{non-quadratic} in general. The \emph{convex} constraint set $\conset\subseteq\real^{\dm_1\times\dm_2}$ is usually a certain norm ball (see Section \ref{section: Application Examples} for examples) playing the role of regularization. This additional constraint is crucial for the success of Problem \eqref{opt: mainProblem} in the following two aspects:
	\begin{itemize}
		\item \emph{Optimization landscape}: Including the constraint $X\in \conset$ leads to a well-behaved landscape of the loss $\loss$: the restricted strong convexity and restricted smoothness conditions (RSC/RSM), or their approximate versions (see Definition \ref{def: relaxedrscrsm}), are satisfied for all the feasible $X$ of \eqref{opt: mainProblem}.  In contrast, in the absence of the constraint $\conset$, RSC/RSM are no longer satisfied. See  
		Section \ref{sec: matrixsensingexponentialfamily} and \ref{sec: matrixCompletionexponentialfamily} for details. 
		\item \emph{Statistical error}: The regularization constraint ensures that the local minimizers of \eqref{opt: mainProblem}
		are nontrivially correlated with the ground truth $\truX$, even when the sample size $n$ is small. In particular, our theoretical results provide the best known statistical recovery guarantees for many RGLM problems. Without this constraint, the solution to  \eqref{opt: mainProblem} could be no better than a trivial constant estimator, as shown in Figure \ref{fig:compare} of Example \ref{example: rscfailureoms}.
	\end{itemize} 
	To leverage the regularization constraint algorithmically, we introduce the regularity projection oracle 
	\begin{equation}
	\begin{aligned}\label{eq: projectionOracle}
	\proj_{r,\conset}(X):\,= \argmin_{\rank(V)\leq r,\,V\in \conset} \fronorm{X -V}
	\end{aligned}
	\end{equation}
	for a given rank parameter $ r>0 $.
	Concrete instances of the oracle are displayed in Section \ref{sec: projection Oracle}. As a primary example, we note that when
	$\conset $ is the Frobenius norm ball, $ \proj_{r,\conset}$ is equivalent to the standard rank-$r$ SVD followed by a projection of the singular values to the Euclidean ball. In this case, $ \proj_{r,\conset}(X) $ computes the best rank-$r$ approximation of $X$ with a \emph{bounded Frobenius norm}. 
	
	\paragraph{Goal and challenges \newcontent{with non-quadratic losses}} 
	Given the regularity projection oracle $\proj_{r,\conset}$, we aim to design an iterative algorithm achieving the following two properties simultaneously:
	\begin{itemize}
		\item Each iteration only requires \emph{one} access to the oracle $\proj_{r,\conset}$, and one computation of the gradient; 
		\item The algorithm converges  to $\truX$ \emph{globally} and \emph{linearly}, with a contraction factor independent of the dimension ($d_1$ and $d_2$), up to \newcontent{a} certain statistical error. 
	\end{itemize}
	To this end, one might be tempted to use a natural projected gradient descent (PG) method: 
	\begin{equation}\label{eq: svp} 
	X_{t+1} = \proj_{r,\conset}\big( X_t -\eta \nabla \loss(X_t) \big) ,\quad t=1,2,\dots
	\end{equation}
	where $\eta>0$ is the step size.
	
	When the constraint is trivial (i.e., $\conset=\real^{\dm_1\times \dm_2}$), the PG algorithm~\eqref{eq: svp} reduces to the well-known Iterative Hard Thresholding (IHT) algorithm~\cite{jain2014iterative, barber2018gradient}, a.k.a.\ singular value projection \cite{jain2010guaranteed}. 
	Existing theory \cite{jain2014iterative,liu2018between} of IHT 
	only applies when RSC/RSM holds for \emph{all} low rank matrices. Unfortunately, this is true only in the simplest settings, such as 
	quadratic loss $\loss$ with Gaussian linear measurements. 
	Even when $\loss$ is quadratic, for harder problems such as matrix completion, RSC/RSM \emph{no longer} holds for all low rank matrices. For such problems, existing theory is scarce and unsatisfactory. While the work by \cite{ding2018leave} proves that IHT does recover $\truX$ in the matrix completion problem, the analysis is complicated and tailored for quadratic loss, and leads to sub-optimal sample complexity bounds. 
	
	For non-quadratic $\loss$ with a general constraint set $\mathcal{C}$, the best applicable result is \cite{barber2018gradient}, which only establishes local convergence from a good initialization.
	Moreover, the quality of initialization is measured by the \emph{concavity parameter} (details in 
	Section \ref{sec: existingWork}) of the feasible region of \eqref{opt: mainProblem}. This concavity parameter is difficult to compute except in the trivial case where $\conset =\real^{\dm_1\times \dm_2}$, and there is no known algorithm guaranteeing a good initialization for \eqref{opt: mainProblem} beyond the quadratic loss setting. Solving the convex relaxation of \eqref{opt: mainProblem} sometimes provides a good initialization, but the computational complexity is prohibiting with a superlinear dependence on the dimension; in particular, known algorithms need  to compute full SVD in each iteration in order to \newcontentyc{simultaneously} enforce the constraint $X \in \conset$ and \newcontentyc{the other constraints of the convex relaxation}.
	
	\paragraph{\newcontentyc{Our} Algorithm and contributions} To achieve the two goals mentioned above, we introduce an algorithm called averaged projected gradient (AVPG), displayed as Algorithm \ref{alg: avg}. Inspired by 
	\cite{allen2017linear}, AVPG is a version of PG~\eqref{eq: svp} after averaging the iterates.  Our contributions henceforth can be summarized as follows: 
	\begin{itemize}
		\item Conceptually, we identify the importance of the regularization constraint $X\in \conset$ and its algorithmic counterpart, the regularity projection oracle.
		\item Algorithmically, we design AVPG based on the regularity projection oracle and show that it converges globally and linearly, and has a low iteration complexity under standard RSC/RSM conditions (and their approximate versions); see Section \ref{sec: algguarantees} and Theorem \ref{thm: mainthm2}.
		\item Statistically, \newcontentyc{we apply AVPG to several RGLMs with non-quadratic losses, such as one-bit matrix sensing/completion, and prove that it recover $\truX$ up to a certain statistical error, which before our work was only achievable by convex relaxation; see Section \ref{section: Application Examples}}.
	\end{itemize} 

\paragraph{Organization} The rest of the paper is organized as follows. In Section \ref{sec: existingWork}, we first review existing approaches to Problem \eqref{opt: mainProblem}, including convex relaxation,
Burer-Monteiro approach, and projected gradient method. We then compare our results to existing approaches and demonstrate situations \newcontentyc{where our approach is advantageous}. In Section \ref{sec: algguarantees}, we present our main algorithm AVPG, discuss the intuition, and establish the theoretical convergence guarantees. In Section \ref{section: Application Examples}, we apply our theoretical guarantees to concrete examples of RGLM and show that AVPG recovers $\truX$ up to \newcontent{a} certain statistical error. \newcontent{We conclude the paper in Section~\ref{sec: conclusion}, where we discuss possible applications of AVPG beyond RGLMs as well as several intriguing questions regarding the gap between theory and practice .}

	\subsection{Related work and comparison} \label{sec: existingWork}
	In this section, we discuss some most related approachs: convex relaxation, projected gradient, and Burer-Monteiro ,
	and why ours is advantageous in certain aspects. 
	To facilitate the discussion, we denote 
	the condition number of $\alpha$-RSC and $\beta$-RSM (see Definition \ref{def: relaxedrscrsm}) as $\kappa =\frac{\beta}{\alpha}$.
	
	\paragraph{Convex relaxation} Convex relaxation usually replaces the rank constraint of Problem \eqref{opt: mainProblem} with some nuclear norm constraint, such as \cite{srebro2005rank, recht2010guaranteed, candes2009exact, negahban2012restricted, lafond2015low, gunasekar2014exponential, lu2015individualized}. 
	We refer readers to \cite[Section 4]{chen2018harnessing} and \cite[Chapter 11]{wainwright2019high} for an overview of this topic. 
	Despite the beautiful theory established, first order algorithm suffers from dimensional number of iterations in theory, and full SVD or at least $\bigO(\dm_1\dm_2)$ operations in fulfilling the constraint set $\conset$.  
	
	\paragraph{Burer-Monteiro approach} Another natural approach is to factor the low-rank matrix as $X=AB$, where $A\in \real^{\dm_1\times r}$ and $B\in \real^{r\times \dm_2}$, then solve Problem \eqref{opt: mainProblem} in variables $A,B$ instead of $X$. This approach was first proposed in \cite{burer2003nonlinear} and recently gained much attention \cite{zheng2016convergence,chen2015fast,ha2018equivalence}. We refer readers to \cite{chen2018harnessing,chi2019nonconvex} for a more comprehensive survey. 
	Algorithms with provably quick convergence \cite{chen2015fast,park2018finding,charisopoulos2019low} typically require an initial solution close (measured by Frobenius norm) to the ground truth $\truX$ or the optimal solution up to a small fraction of the singular value of $\truX$. 
	However, effective and efficient initialization is only available for the \emph{quadratic} loss problems.
	Another line of work studies the the landscape of a penalized or constrained version of \eqref{opt: mainProblem} with the above variables $A,B$ and characterizes when there is no spurious local minimum   \cite{ge2017no,zhu2018global, zhang2018primal}. However, the conditions for such results are quite stringent: either the loss is quadratic, or the condition number $\kappa$ must be very close to one; otherwise, spurious local minima may exist \cite[pp. 3-4]{zhu2018global}. Note that neither of the aforementioned conditions is satisfied for the problem of one bit matrix sensing/completion if the constraint $X \in \conset$ is absent; we discuss this issue in Sections \ref{sec: matrixsensingexponentialfamily} and \ref{sec: matrixCompletionexponentialfamily}. Even with this constraint and for favorable case of one-bit matrix completion, the condition number $\kappa$  may not be close to one, making existing landscape results inapplicable; see Section \ref{sec: furhterdisccussionOnConditionNumber} in the appendix for a detailed discussion.

	\paragraph{Projected gradient method} While the PG method \eqref{eq: svp} sometimes works well empirically, its theoretical guarantees is far from satisfactory, as mentioned earlier. Here we explain in details the ``concavity parameter'' defined in \cite[Equation (5)]{barber2018gradient}, and related convergence guarantees. Denote the set of matrices in $\real^{\dm_1\times \dm_2}$ with rank at most $r$ as $\real^{\dm_1\times\dm_2}_r$. The concavity parameter for the set $\conset\cap \real^{\dm_1\times\dm_2}_r$ at a point $X\in\conset\cap \real^{\dm_1\times\dm_2}_r$ is defined as 
	\yc{Have we defined $\real^{\dm_1\times\dm_2}_r$?}
	\begin{equation*}
	\gamma_{X}(\conset\cap \real^{\dm_1\times \dm_2}_r) := \sup\left \{ \frac{\inprod{Y-X}{Z-X}}{\norm{Z-X}\fronorm{Y-X}^2} \,\big|\, Y\in \conset\cap \real^{\dm_1\times\dm_2}_r , \,Z\,\text{ such that} \;\proj_{r,\conset}(Z)=X \right \}.
	\end{equation*}
	where $\norm{\cdot}$ is some arbitrary norm. The convergence guarantee of PG requires initialization in a neighborhood of $\truX$ and that the condition $\gamma_{X}(\conset\cap \real^{\dm_1\times \dm_2}_r)\norm{\nabla \loss(X)}<\frac{\alpha}{2}$  holds \emph{uniformly} for all $X$ in the neighborhood \cite[Equation (14)]{barber2018gradient}. 
	\newcontent{However, it should be noted that 
	once $r >\rank(\truX)$, the quantity $\gamma_{X}$ is expected to approach $+\infty$ \cite[End of Section 2.1]{barber2018gradient}. 
	In particular, if $\conset = \real^{\dm_1\times \dm_2}$, according to \cite[Lemma 5]{barber2018gradient}, $\gamma_{X}=\infty$ for 
	$r\geq \trurank$,  and hence the main result in \cite[Theorem 3]{barber2018gradient} \emph{does not apply}. On the contrary, our result is still applicable when $r\geq \trurank$. Even assuming the rank parameter $r$ is correctly specified, $r=\trurank$,}
	there is no result for bounding $\gamma_{X}(\conset\cap \real^{\dm_1\times \dm_2}_r)$  except for the trivial case
	$\conset = \real^{\dm_1\times \dm_2}$, and $\norm{\cdot}$ being the operator norm; in particular, there lacks a bound even when 
	$\conset$ is a Frobenius norm ball, let alone an infinity norm ball. We note that there is no simple monotone relation such as
	$\gamma_{X}(\conset\cap \real^{\dm_1\times \dm_2}_r)\leq \gamma_{X}(\real^{\dm_1\times \dm_2}_r)$ as the sets being 
	maximized over does not simply become larger by dropping $\conset$. Hence for interesting constraint set $\conset$, to apply the results
	of \cite{barber2018gradient}, one would need significant additional work to estimate $\gamma_{X}(\conset\cap \real^{\dm_1\times \dm_2}_r)$ and can only \newcontentyz{hope to} guarantee local convergence \newcontent{with the rank parameter $r$ correctly specified.}

	\paragraph{Comparison} For a fair comparison, we consider the case when the projection oracle \eqref{eq: projectionOracle} has comparable (or less) complexity in forming the gradient for low rank matrices. One example is that $\conset$ is a Frobenius norm ball, the corresponding RGLM has Gaussian measurements and the $\loss_{\nsample}$ can be potentially non-quadratic. The projection oracle for this example reduces to $r$-SVD plus some scaling, and can be computed in linear time of matrix vector product of input $X$ \cite{allen2016lazysvd}. 
	As explained earlier, no existing algorithm provably works in this regime, as they 
	either suffer from extraordinary polynomial costs such as the convex relaxation approach based ones, or lack guarantees on convergence such as the IHT approach or the Burer-Monteiro approach based ones. As mentioned earlier, even if the projection oracle \eqref{eq: projectionOracle} is not available, our 
	identification of the projection oracle reveals the key and critical component in solving the rank-constrained generalized linear model.

	\paragraph{Notation} We introduce the shorthand $d= \max\{\dm_1,\dm_2\}$. For a positive integer $n$, the notation $[n]$ stands for $\{1,\dots,n\}$. We equip the linear 
	space $\real^{\dm_1\times \dm_2}$ with the trace inner product: for $A,B\in \real^{\dm_1\times \dm_2}$, $\inprod{A}{B}= \mathbf{tr}(A^\top B) = \sum_{i,j}A_{ij}B_{ij}$. For a given norm $\norm{\cdot}$, $\mathbb{B}_{\norm{\cdot}}(\xi)$ denotes the associated ball centered at origin with radius $\xi>0$. We make use of several matrix norms, including the Frobenius norm $\fronorm{\cdot}$, the operator norm (largest singular value) $\opnorm{\cdot}$, the nuclear norm (sum of singular values) $\nucnorm{\cdot}$, and the infinity norm (maximum absolute value of entries) $\infnorm{\cdot}$.

	\section{Algorithm and guarantees} 
	\label{sec: algguarantees} 
	
	In this section, we present details of the averaged projected gradient algorithm (Section \ref{sec: projection Oracle}) and provide convergence guarantees under an approximate RSC/RSM condition (Section \ref{sec:guarantee}).

	\subsection{Algorithm description}
	\label{sec: projection Oracle}
	Given the regularity projection oracle \eqref{eq: projectionOracle}, AVPG is displayed as Algorithm \ref{alg: avg}. 
	Each iteration of AVPG consists of three steps: (i) a choice of step size, (ii) a projected gradient step, and (iii) an averaging step. 
	Note that the initial iterate is assumed to lie in $\conset$, which is merely for convenience as the projection ensures this property for all future iterates.
	
	\begin{algorithm}[tbh]
		\caption{Averaged projected gradient method (AVPG)}
		\label{alg: avg}
		\begin{algorithmic}
			\STATE {\bfseries Input:} A rank estimate $r$, an initial iterate $X_0 \in \conset \in\real ^{\d_1\times \d_2}$ with $\rank(X)\leq r$, 
			step size parameter $\eta_0\in[0,1]$, RSM parameter estimate $\beta$, a period integer $t_0\in \integer$
			\FOR{$t=1,2,\dots,$ }
			\STATE \textbf{Choice of step size:} if $t$ is an integer multiple of $t_0$, set $\eta =1$, otherwise, $\eta =\eta_0$.   
			\STATE \textbf{Projection step:} $V_t = \proj_{r,\conset}\left(  X_t -\frac{1}{\beta \eta}\nabla \loss(X_t) \right)$
			\STATE \textbf{Averaging step:} $X_{t+1} = (1-\eta) X_t + \eta V_t$.
			\ENDFOR	
		\end{algorithmic}
	\end{algorithm}
	
	\paragraph{The role of step size and period $t_0$} Per our choice of step size, AVPG runs in periods of length $ t_0 $. Within each period, we average the projected solution $ V_t $ with the previous iterate $X_t$; at the end of the period, the iterate is set to $ V_t $ without averaging. By sub-additivity of rank, the rank of the iterate is always bounded by $rt_0$. The boundedness of rank is desirable both computationally and theoretically. Computationally, the boundedness of the rank benefits (i) the storage of the iterate, and (ii) the time in computing the gradient and projection oracle under certain structure of $A_i$. Theoretically, the bounded rank enables us to use the approximate RSC/RSM condition to prove 
    guarantees as those properties are restricted to low rank matrices. \ld{defer the discussion of t0 infinity later}
    
    It is tempting to set $t_0=1$, in which case there is no averaging step and AVPG reduces to PG. 
    However, this choice of $t_0$ destroys the additional leeway provided by the averaging step, which is crucial for establishing theoretical guarantees as we explain next.
    In Theorem \ref{thm: mainthm2}, we require $t_0$ to be on the order $\bigO(\condn\log\condn)$.

	\paragraph{The role of averaging} 
	
	Compared to the naive PG method~\eqref{eq: svp}, AVPG has an additional averaging step. This step is crucial in establishing a linear convergence guarantee (given in Theorem~\ref{thm: mainthm2}) that is valid for a general set $ \conset $, without relying on additional structures of $\conset$. To explain the intuition, let us assume that 
	the objective function $\loss$ is $\alpha$-strongly convex and $\beta$-smooth, i.e., Definition~\ref{def: relaxedrscrsm} with $ \rscerror=\rsmerror=0 $ and $ r=d $ (cf.\ \cite[Definition 2.1.2]{nesterov2013introductory}). With $ \Delta_t := X_t-\truX $, a critical step in our analysis involves the following chain of inequalities: 
	\begin{equation}\label{eq: avpgconvstep1}
	\begin{aligned}
	\loss(X_{t+1})\overset{(a)}{=}\loss((1-\eta)X_t+\eta V_t) & \overset{(b)}{\leq}\loss(X_t)+\eta \inprod{\nabla \loss(X_t)}{V_t-X_t} + \frac{\beta\eta^2}{2}\fronorm{X_t-V_t}^2 \\
	& \overset{(c)}{\leq}\loss(X_t)+\eta \inprod{\nabla \loss(X_t)}{-\Delta_t} + \frac{\beta\eta^2}{2}\fronorm{\Delta_t}^2 ,
	\end{aligned}
	\end{equation}
	where step $(a)$ follows from the definition of $X_{t+1}$, step $(b)$ follows from $\beta$-smoothness, and step $(c)$ follows from the optimality of $V_t$ in the definition~\eqref{eq: projectionOracle} of the projection  $ \proj_{r,\conset} $.
	
	The averaging step allows for additional leeway, provided by $\eta$, in steps $(a)$ and $(b)$. This in turns enables step $ (c) $, which holds without appealing 
	to other properties of the projection oracle. Without averaging, one may replace step $ (b) $ with an application of $\beta$-smoothness \emph{to the two iterates} $X_{t+1}$ and $X_t$, leading to the inequality 
	$
	\loss(X_{t+1})\leq \loss(X_t)+ \inprod{\nabla \loss(X_t)}{X_{t+1}-X_t} + \frac{\beta}{2}\fronorm{X_{t+1}-X_t}^2.
	$
	To proceed at this point, one would need to analyze how the projection interplays with the \emph{difference} between iterates $\fronorm{X_{t+1}-X_t}^2$. Doing so typically requires exploiting the delicate properties of SVD~\cite{jain2014iterative,liu2018between} and specific structures of the set $\conset $, such as the local concavity~\cite{barber2018gradient}. In contrast, our analysis is much simpler, while holds more generally. Such generality allows us to instantiate our convergence guarantee in a diverse range of concrete RGLM problems (see Section~\ref{section: Application Examples}), in which the interplay between SVD and the set $ \conset $ is non-trivial and crucial.

	\paragraph{Computing projection oracle}
	
	In many cases, the oracle $ \proj_{r,\conset} $ can be computed via  rank-$r$ SVD (\newcontent{which gives the best rank-$r$ approximation in terms of Frobenius norm}):
	
	\begin{itemize}
		\item If $\mathcal{C} = \real^{\dm_1\times \dm_2}$, then $ \proj_{r,\conset}(X) $ is given by the rank-$r$ SVD of $ X $. 
		\item If $\mathcal{C} = \mathbb{B}_{\fronorm{\cdot}}(\xi)$ (resp.\ $ \mathbb{B}_{\nucnorm{\cdot}}(\xi) $),  then $ \proj_{r,\conset}(X) $ is given by the rank-$r$ SVD of $X$ followed by a projection of the $r$ singular values to the $\ell_2$ (resp.\ $ \ell_1 $) norm ball in $\real^{r}$ with radius $\xi$.
		\item More generally, if $\mathcal{C}$ is the ball of Schatten-$p$ norm with radius $\xi$, then $ \proj_{r,\conset}(X) $ is given by the 
		rank-$r$ SVD followed by a projection of $r$ singular values to the $\ell_p$ norm ball in $\real^{r}$ with radius $\xi$. This is true even when
		$0<p<1$; see Lemma \ref{lem: projection} for the proof. 
	\end{itemize}
\newcontent{ 
	Note that the rank-$ r $ SVD of $ X $ can be computed using $ m $ matrix-vector product operations of $ X $, with $ m $ being linear in the rank $ r $.\footnote{More precisely, achieving an $ \epsilon $ error requires $ m  = \min \left\{\tilde{\bigO}(\frac{r}{\sqrt{\epsilon}}),\tilde{\bigO}(\frac{1}{\sqrt{\texttt{gap}}}\log \frac{1}{\epsilon})\right\}$ by the results in~\cite{allen2016lazysvd}, where $\texttt{gap} := [\sigma_r(X)-\sigma_{r+1}(X)]/\sigma_r(X)$ is the eigen gap of $ X $. Note that the first term in the expression of $ m $ is independent of $ \texttt{gap}$. Here	$\tilde{\bigO}$ omits logarithmic factors.} In our RGLM example, this means that the SVD can be computed in time linear in the number of the matrix-vector product with the gradient $\nabla \mathcal{L}$. 
	}
	
	There are other interesting choices of $\mathcal{C}$ not defined by the  singular values, e.g., the $ \ell_\infty $ norm ball $ \mathbb{B}_{\infnorm{\cdot}}(\xi)$. In this case,  
	the oracle $  \proj_{r,\conset} $  can be computed by alternating projection~\cite{lewis2014nonsmooth}, which  works well in our experiments (see Appendix~\ref{sec: numerics}), though its convergence property and running time are more involved due to the non-convexity.
	
	\subsection{Convergence guarantee under approximate RSC/RSM}
	\label{sec:guarantee}
	
	To state our convergence guarantees for AVPG, we introduce notions of approximate restricted strong convexity and restricted smoothness. 
	
	\begin{defn}\label{def: relaxedrscrsm}
		The loss function $\loss$ satisfies approximate $(\rscerror,r,\rsc,\conset)$-RSC and $(\rsmerror,r,\rsm,\conset)$-RSM for 
		some $\rscerror, \rsmerror\geq 0$ and convex $\conset$ if for all matrices $X,Y\in \conset$ with rank at most $r$, there hold the inequalities
		\begin{equation}\label{eq: rscrsm}
		\frac{\alpha}{2}\fronorm{X-Y}^2-\rscerror\leq \loss(X)-\loss(Y) - \inprod{\nabla \loss(X)}{Y-X}\leq  \frac{\beta}{2}\fronorm{X-Y}^2+\rsmerror.
		\end{equation}
		If $\rscerror=\rsmerror =0$, we say that $\loss$ satisfies $(r,\rsc,\conset)$-RSC and $(r,\rsm,\conset)$-RSM.
	\end{defn}
	Standard RSC/RSM assumption \cite[Definitions 1 and 2]{jain2014iterative} corresponds to $\rscerror=\rsmerror =0$ and $\conset=\real^{\dm_1\times\dm_2}$. Such a strong assumption is typically required in the analysis of IHT~\cite{jain2010guaranteed,jain2014iterative,liu2018between}. In comparison, our definition allows for the additional constraint set $\conset$ and error terms $\rscerror$ and $\rsmerror$, and is hence less restrictive. In the RGLM setting of interests, these error terms account for the statistical error due to the finite sample size and the measurements structure, 
	and vanishes to $0$ at the rate $\bigO(\frac{\trurank \dm \log\dm  }{n})$ (see Corollary \ref{cor: distancegaussianmearsurement}, \ref{cor: distanceentrywisesampling}, 
	and \ref{cor: distancepairwisesampling}). Moreover, the relaxation of standard RSC/RSM is essential for examples in Section \ref{sec: matrixCompletionexponentialfamily} as discussed in 
	Section \ref{sec: Essentiality of approximate RSC/RSM}.
    The constants $\alpha$ and $\beta$ in the RGLM setting of interests should be dimension independent constants (see Lemma \ref{lem: rscrsmsmallgradientgaussianmearsurement}, \ref{lem: rscrsmsmallgradientEntrywiseSampling}, and \ref{lem: rscrsmSmallGradientPairwiseSampling}), \newcontent{and hence so is the condition number $\condn$.}  
   	Note that the above definition also appears in~\cite{barber2018gradient} for the analysis of projected gradient descent. 
	
	We now state the theoretical guarantees, whose 
	proof is deferred to Appendix \ref{sec:proof_mainthm2}. 
	We introduce the shorthands $\condn:\,= \rsm/\rsc$ for the condition number, $\Delta_t := X_t-\truX$ for the iterate difference to the ground truth, and  $h_t := \loss(X_t)-\loss(\truX)$ for the objective difference.
	
	\begin{thm}\label{thm: mainthm2}
		Suppose $\loss$ satisfies approximate $(\rscerror,rt_0,\rsc,\conset)$-RSC and $(\rsmerror,rt_0,\rsm,\conset)$-RSM
		for $r\geq \trurank$ and $t_0\geq \lceil 4 \kappa \left(\log 4\kappa +1\right) \rceil $.
		Let $\nablasmall := \opnorm{\nabla \loss(\truX)} $, $\eta_0 = \frac{1}{4\kappa}$, and $s$ be the largest 
		integer so that $st_0\leq t$. Also let $\tau_\star = \min _{1\leq \tau \leq t+1}h_{\tau}$, Then the iterate
		$X_t$ from Algorithm \ref{alg: avg} with parameters $\eta_0$, $\beta$, $t_0$, 
		and $r$  satisfies the bounds
		\begin{equation}
		\begin{aligned}\label{eq: thm2eq1} 
		h_{\tau_\star}	\leq \max \left \{ (1-\frac{1}{4\kappa})^{t}(4\kappa)^{s}h_0, \epsilon_{n}\right\} 
		\quad\text{and}\quad 
		\fronorm{\Delta_{\tau_\star}}^2\leq \frac{4}{\alpha} \max \left \{ (1-\frac{1}{4\kappa})^{t}(4\kappa)^{s}h_0, \epsilon_{n}\right\},
		\end{aligned}
		\end{equation}
		where $\epsilon_{n} = \frac{4\condn}{\alpha}( 
		rt_0+\trurank)\nablasmall ^2+  \nablasmall\sqrt{\frac{8t_0r\rscerror}{\rsc}}+ \nablasmall \sqrt{\frac{64t_0r\condn \rsmerror}{\rsc}} +2\condn\rscerror +2\rsmerror$.
	\end{thm}

\paragraph{Interpretation of Theorem \ref{thm: mainthm2}} To better understand the above theorem, let us assume that the AVPG algorithm is run for $ k $ periods, i.e.,  $t= kt_0$ iterations. In this case,
	the bounds \eqref{eq: thm2eq1} become 
	\begin{equation*}
	\min _{1\leq \tau \leq t+1}h_{\tau}  \leq \max \Big \{ {e}^{-k}h_0, \epsilon_{n} \Big\}
	\quad\text{and}\quad 
	\fronorm{\Delta_{\tau_\star}}^2 \leq \frac{4}{\alpha} \max \Big \{e^{-k}h_0, \epsilon_{n}\Big\}. 
	\end{equation*} 
	Each of the above bounds involves two terms. The first term $e^{-k}h_0$ corresponds to the optimization error, which shrinks geometrically at every period, i.e., every $t_0= \lceil 4 \kappa( \log 4\kappa +1) \rceil $ iterations. This geometric convergence holds up to a statistical error given by the second term  $\epsilon_{n}$, which is on the order $\bigO(\frac{\trurank \dm \log \dm }{n})$ for 
	RGLM as shown in Section \ref{section: Application Examples}.
	
\paragraph{Comparison with IHT}	Let us compare our guarantees to those for IHT, the projected gradient method \ref{eq: svp} with $\conset=\real^{\dm_1\times \dm_2}$. Our comparison is only for accuracy 
	$\epsilon >\epsilon_{n}$, as 
	feasible matrices of our problem \eqref{opt: mainProblem} are statistically equally good estimators of $\truX$ once they achieve the error $\epsilon_{n}$.
	The work~\cite{jain2014iterative,liu2018between} shows that the iteration complexity of IHT is 
	$\bigO(\kappa \log (\frac{h_0}{\epsilon}))$, whereas ours is $\bigO(\kappa \log \kappa \log (\frac{h_0}{\epsilon}))$. Note that IHT requires computing the top  
	$\Omega(\kappa ^2 \trurank)$ singular values/vectors in each step to ensure convergence \cite{liu2018between}, while ours only requires the top $\trurank$ ones. Therefore,  to achieve $\epsilon$-accuracy, the total work required by IHT amounts to a  number of
	$\bigO(\kappa^3\trurank\log (\frac{h_0}{\epsilon}) )$ rank-1 SVD computation, whereas AVPG requires $\bigO(\condn \trurank\log \condn \log (\frac{h_0}{\epsilon}))$ and is better than IHT by a factor of $\frac{\condn^2}{\log\condn}$. 
	
\begin{figure}
	\centering
	\includegraphics[width=0.4\linewidth]{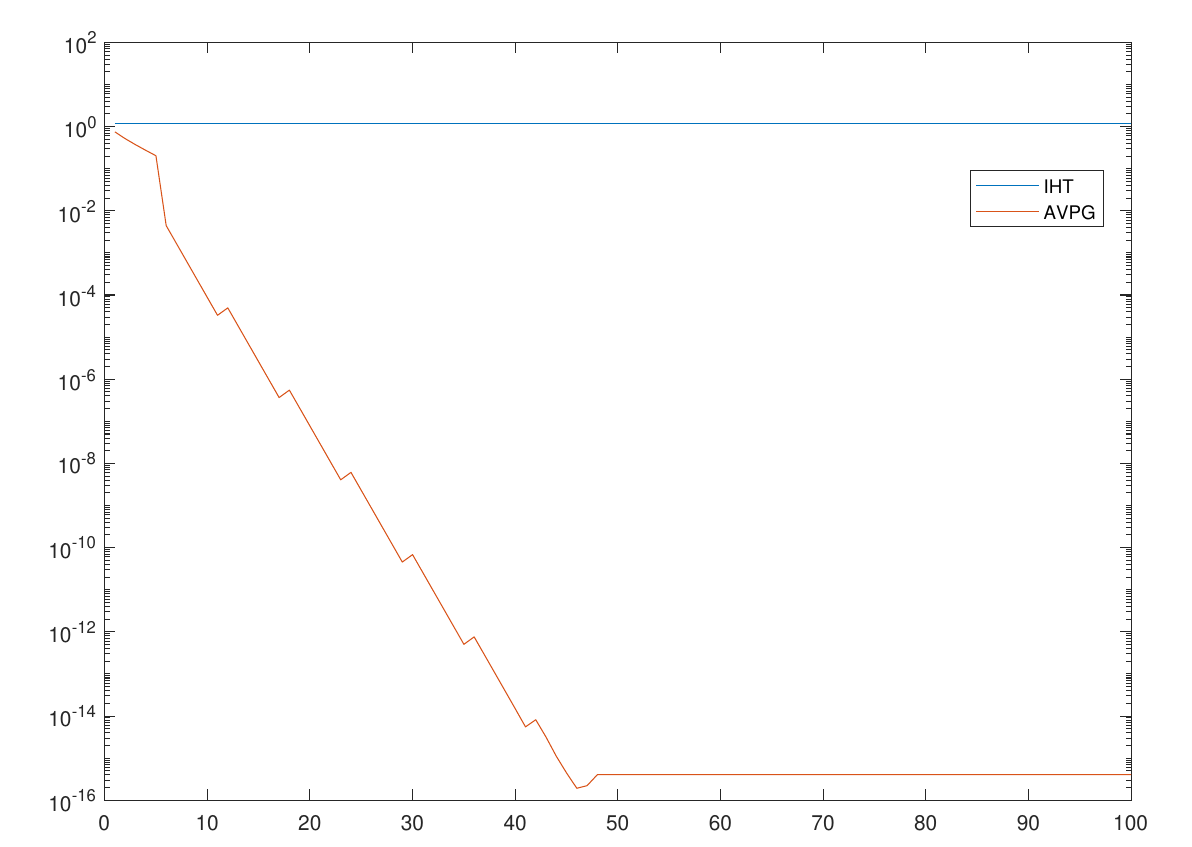}
	\caption{Comparison of IHT and AVPG on the weighted matrix completion problem $\min_{X\in \real^{d\times d}}\fronorm{W\odot (X-\truX)}^2$, where $d=50$, and $W$ is the matrix with all ones except the $(1,1)$-th and $(2,2)$-th entry whose values are $2$, and $\truX$ is a rank $1$ matrix with values all zero except the top left $2\times 2$ block which has entries all $1$. We note the worse dependence on condition number of IHT comparing with AVPG is not an theory artifact but does happen in this example.}
	\label{fig:wmclargermatrixplot}
\end{figure}

	\newcontent{The per iteration dependence on condition number is not an artifact of the theory. In Figure \ref{fig:wmclargermatrixplot}, we show the results of IHT and AVPG applied to the weighted matrix recovery problem on \cite[pp 3-4]{zhu2018global}. Specifically, we consider $\min_{X\in \real^{d\times d},\;\rank(X)\leq r}\fronorm{W\odot (X-\truX)}^2$, where $d=50$, and $W$ is the matrix with all ones except the $(1,1)$-th and $(2,2)$-th entry whose values are $2$, and $\truX$ is a rank $1$ matrix with all zero  values except for the top left $2\times 2$ block which has all one entries. We start the algorithm at a point $M_0$ with  all zero  values except for the top left $2\times 2$ block which is $\frac{3}{5} \begin{bmatrix}
			1 & -1 \\ -1 & 1
		\end{bmatrix}$,  and with the choice of rank $r=1=\trurank(\truX)$. Here $\odot$ denotes the Hadamard product. The stepsize of IHT  is chosen to be $\bigO(1/\beta)$ as suggested by most theory papers  \cite{jain2014iterative,liu2018between}. The stepsize of AVPG is chosen to be $\eta_0 =\bigO(\frac{1}{\condn})$ as suggested by our theory. This problem has condition number $4$. 		
		Note that IHT stays at the starting point $M_0$, meaning that $M_0$ is a non-optimal fixed point of 
	iteration \eqref{eq: svp}, while AVPG is able to move away from this starting point.}

	\begin{table}
		\centering 
	\begin{tabular}{c|c|c|c}
		Algorithm & Applicability & Per iteration complexity &  Iteration complexity\\
		\hline
		IHT &  $\conset = \real^{\dm_1\times \dm_2}$ &  $\Theta(\kappa ^2 \trurank)$ singular values/vectors &$\bigO\left(\kappa \log (\frac{h_0}{\epsilon})\right)$ \\
		\hline 
		AVPG & Any convex $\conset$ & $\trurank$ singular values/vectors& $\bigO(\kappa \log \kappa \log (\frac{h_0}{\epsilon}))$\\
	\end{tabular}
\caption{Comparison of IHT and AVPG.}
    \end{table} 

	\section{Consequences for solving RGLM} \label{section: Application Examples}
	
	In this section, we apply our general Theorem~\ref{thm: mainthm2} to concrete settings of RGLM, by calculating the parameters of the approximate RSC/RSM and the gradient norm $\opnorm{\nabla \loss_{\nsample}(\truX)}$. The results in this section explain why
	the regularity constraint $\conset$ is crucial as claimed in the introduction, highlighting
	the key role of the projection oracle \eqref{eq: projectionOracle}. 
	We consider several examples based on the form of the measurement matrices $\{A_i\}$: including
	Gaussian measurements (Section \ref{sec: matrixsensingexponentialfamily}), entrywise sampling for matrix completion (Section \ref{sec: matrixCompletionexponentialfamily}), and pairwise sampling for rank aggregation (Section \ref{sec: pairwisesampling}).
	
	\subsection{Gaussian measurements and Frobenius norm ball}\label{sec: matrixsensingexponentialfamily}
	Let us first explain the setup of matrix sensing and one-bit matrix sensing.
	\subsubsection{Problem setup} 
	Suppose that the measurement matrices $\{A_i, i=1,\dots,\nsample \}$ are independent of each other and 
	have i.i.d.\ standard Gaussian entries. For the distribution of the response $\{y_i\}$ in the  RGLM~\eqref{eq: glm}, we are interested in the following two settings:
	\begin{enumerate}
		\item \emph{Matrix sensing}: $\psi(\theta) = \frac{1}{2}\theta^2$, and $c(\sigma) = \sigma^2$. In this case, the distribution of $y_i$ is Gaussian with mean 
		$ \inprod{A_i}{\truX}$ and variance $\sigma^2$. 
		\item \emph{One-bit matrix sensing}:  $\psi(\theta) = \log(1+\exp(\theta))$, and $c(\sigma) = 1$. The distribution of $y_i$ is Bernoulli with probability $\frac{\exp(\inprod{A_i}{\truX})}{1+\exp(\inprod{A_i}{\truX})}$, which is a logistic function of $\inprod{A_i}{\truX}$.
	\end{enumerate}  
	In words, in matrix sensing  $y_i$ is the linear measurement $\inprod{A_i}{\truX}$ corrupted by additive Gaussian noise, whereas in one-bit matrix sensing, $y_i$ contains only binary information of  $\inprod{A_i}{\truX}$.
	
	Next, we explain the choice of $\conset$ and why such choice is critical for the successful recovery of $\truX$.
	
	\subsubsection{The choice of $\conset$ and its importance} For Gaussian $ \{A_i\} $, the operator $\Amap:\real^{\dm_1\times \dm_2} \to \real^{\nsample}$ defined by $[\Amap(X)]_{i}=\inprod{A_i}{X}$,  
	satisfies the Restricted Isometric Property (RIP); i.e., 
	$
	(1-\frac{1}{16})\fronorm{X}\leq \twonorm{\Amap(X)}\leq (1+\frac{1}{16})\fronorm{X}
	$
	for any rank-$r$ matrix $X$, with high probability provided that $ \nsample $ is sufficiently large~\cite{candes2011tight}. 
	Accordingly, we choose the regularization constraint $\conset$ to be the Frobenius norm ball $\mathbb{B}_{\fronorm{\cdot}}(\cdot)$. Below we discuss this choice and explain why it is crucial for the \emph{non-quadratic} loss associated with one-bit matrix sensing. 
	
\begin{figure}[tb]
		\begin{subfigure}[(a)]{1\textwidth}
			\centering 
			\includegraphics[width=1\linewidth]{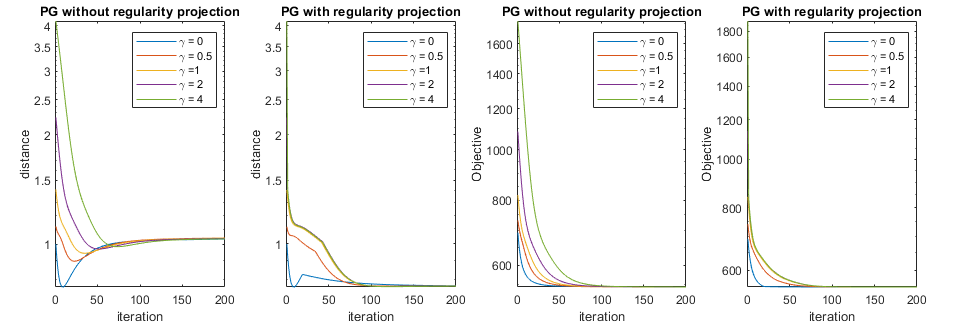}
			\caption{Comparison of relative distance to ground truth $\frac{\fronorm{X_t - \truX}}{\fronorm{\truX}}$, 
			and objective  $ \loss(X_t) $ for PG, projected gradient \eqref{eq: svp}.}
			\label{fig:figure_dits_log}
		\end{subfigure}%
		
		\begin{subfigure}[(b)]{1\textwidth}
			\centering 
			\includegraphics[width=1\linewidth]{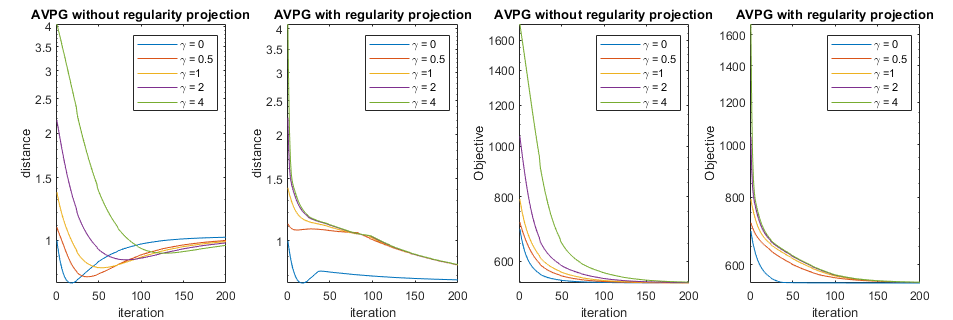}
			\caption{Comparison of relative distance to ground truth $\frac{\fronorm{X_t - \truX}}{\fronorm{\truX}}$, 
				and objective  $ \loss(X_t) $ for AVPG.}
			\label{fig:figure_obj_log}
		\end{subfigure}
		\caption{Comparison of projected gradient and AVPG with and without the regularity projection  oracle (i.e.,  with projection $\proj_{r,\mathbb{B}_{\fronorm{\cdot}}(1)}$ and with $\proj_{r,\real^{\dm_1\times\dm_2}}$, respectively).}
		\label{fig:compare}
	\end{figure}

	The Hessian of the loss function $\loss_{\nsample}$ is given by 
	$\nabla ^2 \loss_{\nsample}(X)[\Delta,\Delta]= \frac{1}{2\nsample} \sum_{i=1}^{\nsample} \psi''(\inprod{X}{A_i}) \inprod{\Delta}{\Asense_i}^2.$     
	For matrix sensing, for which $ \loss_{\nsample} $ is quadratic, we have  $\psi''=1$, a constant regardless of $X$. For one bit matrix sensing, on the other hand,
	we have $\psi''(\theta)\rightarrow 0$ for $\theta \rightarrow \pm \infty$. In this case, the condition number of  $\nabla ^2 \loss_{\nsample}(X) $ is unbounded 
	if we consider all low-rank matrices. Restricting to matrices in the Frobenius norm ball $\conset$ ensures a bounded condition number, so that $ \loss_{\nsample} $ is well behaved due to RIP. Below we corroborate the above arguments by a numerical example.
	\begin{exmp}\label{example: rscfailureoms}
		We generate a random rank-1 matrix 
		$\truX$ with $\fronorm{\truX}=1$, and sample $1000$ data points $(y_i,A_i)$ using the one-bit matrix sensing model. We then apply projected gradient \eqref{eq: svp}, as well as AVPG, using the projection $\proj_{r,\mathbb{B}_{\fronorm{\cdot}}(1)}$ or $\proj_{r,\real^{\dm_1\times\dm_2}}$, i.e., with or without the regularity oracle. We consider random initialization with different Frobenius norm $\fronorm{X_0}=\gamma$ for $\gamma =0,0.5,1,2,4$. The distance to ground truth and objective value of the iterates are shown in Figure~\ref{fig:compare}.
	\end{exmp}
	
	\newcontent{As we can see, all the  methods converge to a stationary point or local minimizer, as the objective value keeps decreasing and 
	approaches stagnancy. However, the regularity projection approaches converge to a better solution, with distance to ground truth approaching $0.77$; the other approaches produce solutions  worse than the trivial estimator $X=0$.}

	Finally, we provide performance guarantees for AVPG applied to matrix sensing and one-bit matrix sensing.
	
	\subsubsection{Theoretical guarantees} 
	Let us first state the following lemma establishing the desired structural properties of the loss $ \loss $, including RSC/RSM (proved in Appendix~\ref{sec: proofOfrscrsmsmallgradientgaussianmearsurement}) and bounds on the gradient (proved in Lemma \ref{lem: gaussiangradienNorm}  in the Appendix). 
	
	\begin{lem}
		\label{lem: rscrsmsmallgradientgaussianmearsurement}
		Suppose $\mathcal{C}= \mathbb{B}_{\fronorm{\cdot}}(\tau \fronorm{\truX})$ for some $\tau \geq 1$, and the measurements $A_i,i=1,\dots,\nsample$ 
		are standard Gaussian. Then there are universal constants $c,C,c_0,c_1>0$ such 
		that if $n\geq c(\lceil\kappa \log \kappa\rceil +1)\trurank \dm$, 	with probability at least $1-\exp(-c_1\dm)$, the following two statements hold:
		\begin{itemize}[noitemsep,topsep=0pt]
			\item (RSC/RSM) the loss function $\loss$ satisfies 
			$(t_0\trurank, \frac{15}{16}\underline{B}, \mathcal{C})$-RSC and   $(t_0\trurank, \frac{17}{16}\overline{B}, \mathcal{C})$-RSM;
			\item (Gradient bound) $\opnorm{\nabla \loss(\truX)}\leq C\sqrt{c(\sigma)\overline{B}} \sqrt{\frac{\dm}{\nsample}}$,
		\end{itemize} 
		where $t_0 = \lceil 4\kappa \left( \log 4\kappa +1\right)\rceil$ and 
		$\kappa = 1.1 \overline{B}/\underline{B}$ where 
		$\overline{B} = \infnorm{\psi''}:\,= \sup_{x\in \real}|\psi''(x)|$, and $\underline{B} = \inf_{|x|\leq \sqrt{\frac{2.4}{c_0}} \tau \fronorm{\truX}} \psi''(x)$.	
	\end{lem}

	\paragraph{Scaling of Lemma \ref{lem: rscrsmsmallgradientgaussianmearsurement}}
	To better understand Lemma \ref{lem: rscrsmsmallgradientgaussianmearsurement}, consider the scenario where the ground truth is a constant, i.e., $\fronorm{\truX}=\bigO(1)$. Such requirement is necessary for constant conditioning of 
	non-quadratic loss due to the non-constancy of $\psi''$. 
	We have $\kappa$ being a universal constant for both cases. For some universal constants 
	$c_1,c_2,c_3,C$, the loss $\loss$ satisfies $(c_1\trurank, c_2, \mathcal{C})$-RSC and   $(c_1\trurank, c_3 , \mathcal{C})$-RSM, and 
	$\opnorm{\nabla \loss(\truX)}\leq C\sqrt{\frac{\dm c(\sigma)}{\nsample}}$ where $c(\sigma)=\sigma^2$ for matrix sensing and $c(\sigma)\equiv 1$ for 
	one-bit matrix sensing. Note that for matrix sensing, the requirement on constant $\fronorm{\truX}$ is not needed, as $\psi''$ is constant.
	
	With Lemma~\ref{lem: rscrsmsmallgradientgaussianmearsurement}, we can bound the distance to $\truX$ by invoking the general Theorem \ref{thm: mainthm2}.
	We assume the input to AVPG is $r=\trurank$, $\beta = \frac{17}{16}\overline{B}$,  $\eta_0 = \frac{1}{4\kappa}$, and $t_0 =\lceil 4\condn( \log 4\condn +1)\rceil$. 
	
	\begin{cor}
		\label{cor: distancegaussianmearsurement}
		Instate the assumptions and notation in Lemma \ref{lem: rscrsmsmallgradientgaussianmearsurement}, and assume 
		AVPG uses the input described above.
		Define $\tau_\star= \arg\min_{1\leq \tau \leq t+1} \loss(X_t)$. Then for some universal constatn $c,c_1$,  with probability at least $1-\exp(-c_1\dm)$, there holds the inequality 
		\begin{equation}\label{eq: boundOfdistancegaussianmearsurement}
		\fronorm{X_{\tau_\star} - \truX}^2 \leq  c{\underline{B}}^{-1}\max\left\{(1-\frac{1}{4\kappa})^{t}(4\kappa)^{s}h_0,{\condn^3} \trurank \left(\log\kappa\right) c(\sigma)\frac{\dm}{\nsample}\right\}, 
		\end{equation}
		where $s$ is the largest integer so that $st_0\leq t$. 
	\end{cor}
	
		\paragraph{Interpretation of Corollary \ref{cor: distancegaussianmearsurement}} To better interpret Corollary \ref{cor: distancegaussianmearsurement}, we let $ t \to \infty $, in which case the second RHS term in~\eqref{eq: boundOfdistancegaussianmearsurement} dominates and corresponds to the statistical error. For matrix sensing, we have $c(\sigma)=\sigma^2$ and hence the error is $\bigO\left( \frac{\sigma^2 \trurank\dm }{\nsample}\right)$. For one-bit matrix sensing, where $ c(\sigma) \equiv 1 $, the error is $\bigO\left(\frac{\trurank \dm}{\nsample}\right)$ when the Frobenius norm of the ground truth is a
		universal constant.
	The second bound is new for non-convex methods, and matches the error bound achieved using the (more computationally expensive) convex relaxation approach ~\cite[Corollary 10.10]{wainwright2019high}.

	\subsection{Entrywise sampling and infinity norm ball}\label{sec: matrixCompletionexponentialfamily}
	Let us first explain the setup of matrix completion and one-bit matrix completion.
	\subsubsection{Problem setup} 
	Let $e_i$ denote the $i$-th standard basis vector in some appropriate dimension. 
	Entrywsie sampling involves 
	measurement matrices  of the form $A_i = \sqrt{\dm_1\dm_2} e_{k(i)} e_{l(i)}^\top $.  

	Here for each $i\in [\nsample]$, the index pair $(k(i),l(i))$ 
	is uniformly sampled from $[\dm_1]\times [\dm_2]$ and independent of anything else.
	We consider the following two settings:
	\begin{enumerate}
		\item \emph{Matrix completion}: $\psi(\theta) = \frac{1}{2}\theta^2$, and $c(\sigma) = \sigma^2$.
		\item \emph{One-bit Matrix completion}:  $\psi(\theta) = \log(1+\exp(\theta))$, and $c(\sigma) = 1$.
	\end{enumerate}
	Analogous to the models in Section \ref{sec: matrixsensingexponentialfamily},  matrix completion corresponds to a partial observation of matrix entries with Gaussian noise, and one-bit matrix completion corresponds to a binary observation.
	
	Next we explain the choice of $\conset$ and its importance.
	\subsubsection{The choice of $\conset$ and its importance}
	Here the regularity constraint is taken to be the $ \ell_\infty $ norm ball, $ \conset = \mathbb{B}_{\infnorm{\cdot}}(\xi)$. This constraint ensures that the matrix is incoherent/non-spiky, well known to be necessary for matrix completion. For the one-bit setting the constraint $ \conset $ is even more crucial, without which the condition number becomes unbounded for reasons similar to before---see Appendix \ref{sec: furhterdisccussionOnConditionNumber} for further discussion. Within~$ \conset $, Lemma~\ref{lem: rscrsmsmallgradientEntrywiseSampling} below shows that the desired approximate RSC/RSM properties hold, which in turn allows us to establish the statistical guarantees in Corollary \ref{cor: distanceentrywisesampling}.
	
	An intriguing question is whether the constraint $\conset$ should be imposed \emph{explicitly} in practice. In previous work, this constraint is sometimes ignored in the experiments \cite{lu2015individualized, davenport20141} and relegated as an 
	artifact of analysis \cite[pp.9]{kallus2020dynamic}.
	For problems with quadratic loss, the work in  \cite{chen2019noisy,ma2019implicit,ding2018leave} proves that the iterates of the algorithm stay in $\conset$ automatically, though their sample complexity requirement is substantially larger than optimal. 
	
	Here we argue for explicit enforcement of $ \conset $. Our experiment result for projected gradient \eqref{eq: svp} and AVPG in Figure \ref{figure: onebitmcexperiment}, whose setting presented in details in Appendix~\ref{sec: numerics}, shows that doing so is very beneficial, especially \emph{when the sample size is limited} and when \emph{the loss is non-quadratic} as in one-bit matrix completion. Imposing $ \conset $  enhances algorithm stability and reduces statistical errors. 
	Without $ \conset $, the estimation error is sometimes worse than a trivial constant estimator, a similar situation as in Example \ref{example: rscfailureoms}. Indeed,
	all the iterates converge to some stationary point or local minimizer, as the objective value keeps decreasing and 
	approaches stagnancy. As mentioned, the distances of the iterates with regularity projection approach $0.39$ while the others approaching a number larger than $1$ (not shown here), worse than the trivial estimator $0$. 
	\footnote{\newcontent{Note that if one is willing to early stop the algorithm, then the generalization error in terms of the distance to the ground truth $\truX$ is actually better. The distance for PG and AVPG without the the regularity projection gets very close to $\truX$ in the beginning. However, determining the stopping time is a nontrivial issue.}
}
	\begin{figure}[tb]
	\begin{subfigure}[(a)]{1\textwidth}
		\centering 
		\includegraphics[width=1\linewidth]{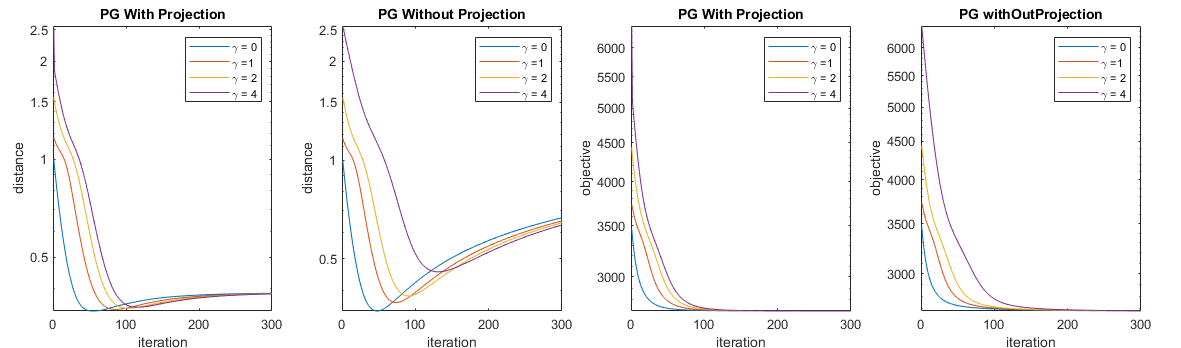}
		\caption{Comparison of relative distance to ground truth $\frac{\fronorm{X_t - \truX}}{\fronorm{\truX}}$, 
			and objective  $ \loss(X_t) $ for PG, projected gradient \eqref{eq: svp}.}
		\label{fig:figure_dits_log_onebit_MC_SVP}
	\end{subfigure}%
	
	\begin{subfigure}[(b)]{1\textwidth}
		\centering 
		\includegraphics[width=1\linewidth]{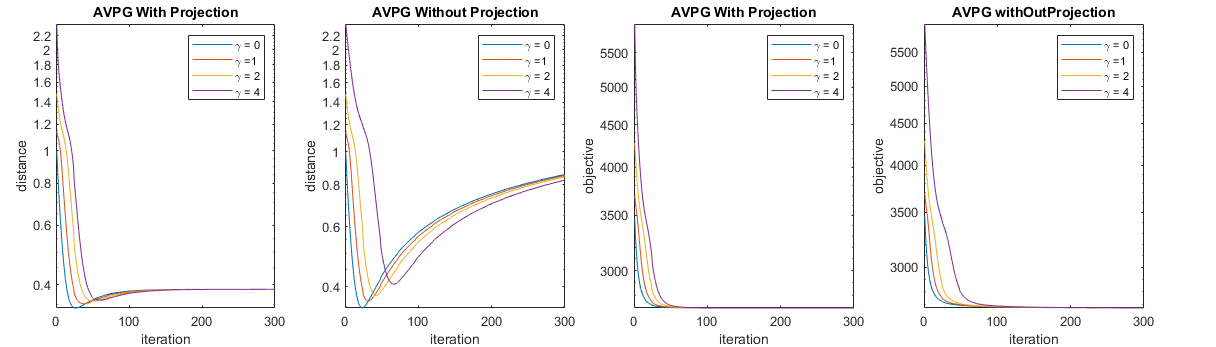}
		\caption{Comparison of relative distance to ground truth $\frac{\fronorm{X_t - \truX}}{\fronorm{\truX}}$, 
			and objective  $ \loss(X_t) $ for AVPG.}
		\label{fig:figure_obj_log_onebit_MC_AVPG}
	\end{subfigure}
	\caption{Comparison of projected gradient and AVPG with and without the regularity projection  oracle (i.e.,  with projection $\proj_{r,\mathbb{B}_{\fronorm{\cdot}}(1)}$ and with $\proj_{r,\real^{\dm_1\times\dm_2}}$, respectively)
		for the one-bit matrix completion problem. The horizontal}
	\label{figure: onebitmcexperiment}
\end{figure}
	
	Finally, we provide theoretical guarantees.
	\subsubsection{Theoretical guarantees} 
	
	As is standard in the matrix completion literature, we introduce the following spikeness 
	measure $\alpha_{\tiny \mbox{sp},\natural} :\,= \frac{\sqrt{\dm_1\dm_2} \infnorm{\truX}}{\fronorm{\truX}}$  of the true matrix $\truX$.
	The following lemma verifies the desired structural properties of the loss $ \loss $, including approximate RSC/RSM  (proved in 
	Appendix~\ref{sec: proofOfrscrsmsmallgradientEntrywisemearsurement}) and bounds on the gradient (proved in Lemma~\ref{lem: smallnormGradientpairEntrywise} in the appendix).
	\begin{lem}
		\label{lem: rscrsmsmallgradientEntrywiseSampling}
		Consider the RGLM 
		with matrix completion or one-bit matrix completion setting. Let the constraint set $\mathcal{C}= \mathbb{B}_{\infnorm{\cdot}}(\frac{\alpha'}{\sqrt{\dm_1\dm_2}}\fronorm{\truX})$ 
		for some $\alpha'\geq \alpha_{\tiny \mbox{sp}}(\truX)$.
		Then there are universal constants $c,\,C,\,c_0,\,c_1,c_2>0$ such 
		that for any $n\geq c\dm \trurank \log \dm$, with probability at least $1-\exp(-c_1\nsample)-c_2d^{-2}$, the following two statements hold:
		\begin{enumerate}[noitemsep,topsep=0pt]
			\item (RSC/RSM)  $\loss$ satisfies $(c_0 \underline{B}\alpha'^2\epsilon_{n},t_0\trurank, \frac{15}{16}\underline{B}, \mathcal{C})$-RSC
			and   $(c_0 \overline{B} \alpha'^2\epsilon_{n},t_0\trurank, \frac{17}{16}\overline{B}, \mathcal{C})$-RSM; 
			\item (Gradient bound)  $\opnorm{\nabla \loss(\truX)}\leq C\sqrt{c(\sigma)\overline{B}\frac{\dm \log \dm}{\nsample}}$. 
		\end{enumerate} 
		Here $\epsilon_{n}=\frac{\condn\log \condn \trurank \dm \log \dm}{\nsample}$, $t_0  = \lceil 4\kappa \left(\log 4\kappa +1 \right)\rceil$, and $\kappa = 1.1 {\overline{B}}/{\underline{B}}$, where 
		$\underline{B}:\,= \inf_{|x|\leq  \alpha'\fronorm{\truX}}\psi''(x)$ and $\overline{B} :\,= \sup_{|x|\leq \alpha'\fronorm{\truX}} \psi''(x)$.
	\end{lem}
	\paragraph{Scaling in Lemma \ref{lem: rscrsmsmallgradientEntrywiseSampling}}
To better understand the scaling in Lemma \ref{lem: rscrsmsmallgradientEntrywiseSampling},  consider the scenario $\alpha'=\alpha_{\tiny \mbox{sp},\natural} $ and 
$\fronorm{\truX}$ are both universal constants, then $\kappa,\overline{B}$, and $\underline{B}$ are also some universal constants. For some universal $c_1,c_2,c_3,c_4,C>0$, the loss $\loss$ satisfies  
$(c_1 \frac{\trurank \dm \log \dm}{\nsample}, c_2\trurank,c_3, \mathcal{C})$-RSC
and   $(c_1 \frac{\trurank \dm \log \dm}{\nsample}, c_2\trurank,c_4, \mathcal{C})$-RSM; and the gradient $\opnorm{\nabla \loss(\truX)}\leq C\sqrt{c(\sigma)\frac{\dm \log \dm}{\nsample}}$ 
		where $c(\sigma)=\sigma^2$ for matrix completion and $c(\sigma)\equiv 1$ for one-bit matrix completion.
	
	Suppose that the input of AVPG is $r=\trurank, \beta = \frac{17}{16}\overline{B}$,  $\eta_0 = \frac{1}{4\kappa}$, and $t_0 =\lceil 4\condn( \log 4\condn +1)\rceil$. The following corollary is immediate from combining Theorem \ref{thm: mainthm2} and Lemma \ref{lem: rscrsmsmallgradientEntrywiseSampling}.
	
	\begin{cor}
		\label{cor: distanceentrywisesampling}
		Instate the assumptions and notation in Lemma \ref{lem: rscrsmsmallgradientEntrywiseSampling}, and suppose AVPG 
		uses the input described above.
		Define $\tau_\star= \arg\min_{1\leq \tau \leq t+1} \loss(X_t)$. Then there exist 
		some universal constatn $c,c'$ such that for any $n\geq c\trurank \dm$,  with probability at least $1-c'd^{-2}$
		\begin{equation}\label{eq: boundcordistanceentrywisesampling}
		\fronorm{X_{\tau_\star} - \truX}^2 \leq  \frac{c}{\underline{B}}\max\left\{(1-\frac{1}{4\kappa})^{t}(4\kappa)^{s}h_0,{\condn} \log\kappa (\condn^2c(\sigma)+\alpha '\sqrt{c(\sigma)\overline{B}}\kappa+(\alpha')^2\overline{B})\frac{\trurank \dm\log \dm}{\nsample}\right\}.
		\end{equation}
		Here $s$ is the largest integer so that $st_0\leq t$.
	\end{cor}
	\paragraph{Interpretation of Corollary \ref{cor: distanceentrywisesampling}} Recall that $c(\sigma)=\sigma$ for matrix completion and  $c(\sigma)\equiv 1$ for one-bit matrix completion.  To better interpret Corollary \ref{cor: distanceentrywisesampling}, we let $t\rightarrow\infty$ and focus on the second RHS term of statistical error in the bound~\eqref{eq: boundcordistanceentrywisesampling}\footnote{\newcontent{Since the first RHS term is geometrically decreasing, we actually only need $t = \bigO(\log (\text{second RHS term}))$ to balance the two term.}}. Also assume that we take $\alpha' = \alpha_{\tiny \mbox{sp},\natural}$ in AVPG. For matrix completion, the statistical error is 	$\bigO((\sigma^2 +\alpha_{\tiny \mbox{sp},\natural}^2 )\frac{\trurank \dm\log \dm }{\nsample})$. For one-bit matrix completion, further assume that $\alpha_{\tiny \mbox{sp},\natural} = \bigO(1)$ and $\fronorm{\truX} =\bigO(1)$, then we have the statistical error bound $\bigO(\frac{\trurank \dm\log \dm}{\nsample})$.
	Both bounds match the those achieved by convex relaxation methods; cf. \cite[Corollary 10.18]{wainwright2019high}.

	\subsubsection{Essentiality of approximate RSC/RSM} \label{sec: Essentiality of approximate RSC/RSM}	
	\newcontent{Here we would like to point out that it is essential to consider the approximate version instead of the standard version $\epsilon_\alpha = \epsilon_\beta =0$ for entrywise sampling. To see this, suppose $d_1=d_2=d$ for simplicity and consider the matrix $X=0$ and $Y = \frac{1}{d} e_1e_1^\top$, a matrix with all zero entries  except the top left entry being $1/d$. Then the standard RSC/RSM fails for the scaling described above.   But the approximate version does hold still. Indeed, for both matrix completion and one-bit matrix completion, the middle term in the RSC/RSM condition in \eqref{def: relaxedrscrsm} is $\loss(X)-\loss(Y) -\inprod{\nabla \loss (X)}{Y-X} = 0$ with high probability whenever $n = \Theta(\trurank\dm \log\dm)$. However, the RSC term, the left term of \eqref{eq: rscrsm}, $\frac{\alpha}{2}\fronorm{X-Y}^2 -\epsilon_\alpha$ is nonzero for $\epsilon_\alpha =0$. Hence the strict RSC cannot hold in this case. However, if we allow  $\epsilon_\alpha = C\frac{\trurank\dm \log\dm}{n}$ for some numerical constant $C$ (recall the scaling in the last paragraph), then $\frac{\alpha}{2}\fronorm{X-Y}^2 -\epsilon_\alpha<0$ and our approximate RSC does hold still. The reason why the approximate RSC/RSM is enough for our purpose is that we shall choose $X = X_t$ and $Y=\truX$ in our analysis. And we only need to consider the case when $\fronorm{X_t-\truX}$ is larger than the statistical error
		(measured by the Frobenius norm).}
	
	\subsection{Pairwise sampling and  infinity norm ball}\label{sec: pairwisesampling}
	In this section, we consider individualized rank aggregation (IRA) setting studied in \cite{lu2015individualized}. 
	
	\subsubsection{Problem setup} The measurement matrix $A_i$ in this setting satisfies that 	$A_i = \sqrt{\dm_1\dm_2} e_{k(i)}(e_{l(i)} -e_{j(i)})^\top$. 
	Here for each $i\in [\nsample]$, the number $k(i)\in [\dm_1]$ is uniformly distributed on $[\dm_1]$ independent of anything else, and  
	$(l(i),j(i))$ is uniformly distributed over $[\dm_2]^2$ independent of anything else. We call 
	such sampling "pairwise" because it always picks two entries in the same row as a pair. 
	The response $y_i$ is Bernoulli, meaning that $\psi(\theta) = \log (1+\exp(\theta))$ and $c(\sigma)\equiv 1$. 

	This model can be considered as users' responses when giving a pair of items in a recommendation system. 
	Each row of $\truX$ 
	represents a user's score for different items. In each sample $(y_i, {k(i)},{l(i)},j(i))$, the $k(i)$-th user
	gives a response $y(i)$,  representing whether she prefers item $l(i)$ to item $j(i)$. The value $y(i)=1$ means that
	she prefers $l(i)$-th item to $j(i)$-th term, otherwise, she prefers the other way. Let us now introduce the 
	constraint set $\conset$.
	
	\subsubsection{The choice of $\conset$} For pairwise sampling, apart from the infinity norm ball (which is 
	imposed for similar reasons of matrix completion)
	the constraint set in $\mathcal{C}$ has an additional constraint that $\mathbb{F}\,:=\{X\mid \sum_{1\leq l\leq \dm_1}X_{kl}=0,\,\text{for all}\, 1\leq k\leq \dm_1\}$
	compared to entrywise sampling. This constraint eliminates identification issue due  to the 
	difference in the measurements $A_i$ and the modeling of the probability that 
	$y_i = 1$, see \cite[Section 2.1]{lu2015individualized} for more information on this condition. 
	Finally, we provide the theoretical guarantees.

	\subsubsection{Theoretical guarantees} The proof of RSC/RSM condition can be found in Section \ref{sec: proofOfrscrsmsmallgradientPairwisemearsurement} in the appendix. The gradient norm condition is proved in Lemma \ref{lem: smallnormGradientpairEntrywise} in the appendix. We 
	summarize the two in the following lemma. The scaling of the parameters of approximate RSC/RSM and the bound of $\opnorm{\nabla \loss (\truX)}$ 
	under the condition, constant $\alpha'=\alpha_{\tiny \mbox{sp},\natural} $ and 
	$\fronorm{\truX}$, follow the same behavior as those for one-bit matrix completion.
	\begin{lem}[RSC/RSM and small gradient for pairwise measurement]\label{lem: rscrsmSmallGradientPairwiseSampling}
		Consider the RGLM 
		with individualized rank aggregation setting. Let the constraint set $\mathcal{C}=\{X\mid \sum_{1\leq l\leq \dm_1}X_{kl}=0,\,\text{for all}\, 1\leq k\leq \dm_1\}\cap  \mathbb{B}_{\infnorm{\cdot}}(\frac{\alpha'}{\sqrt{\dm_1\dm_2}}\fronorm{\truX})$ for some $\alpha'\geq \alpha_{\tiny \mbox{sp}}(\truX)$.
		Then there is a universal constant $c,\,C,\,c_0,\,c_1,c_2>0$ such 
		that for any $n\geq c\trurank \dm \log \dm$, with probability at least $1-\exp(-c_1\nsample)-c_2d^{-2}$, the following two hold. 
		\begin{enumerate}[noitemsep,topsep=0pt]
			\item The loss function $\loss$ satisfies $(c_0 \underline{B}\alpha'^2\epsilon_{n},t_0\trurank, \frac{31}{16}\underline{B}, \mathcal{C})$-RSC 
			and   $(c_0 \overline{B} \alpha'^2\epsilon_{n},t_0\trurank, \frac{33}{16}\overline{B}, \mathcal{C})$-RSM 
			\item The gradient satisfies that $\opnorm{\nabla \loss(\truX)}\leq C\sqrt{c(\sigma)\overline{B}\frac{\dm \log \dm}{\nsample}}$.
		\end{enumerate} 
		Here $\epsilon_{n}=\frac{\condn\log \condn \trurank \dm \log \dm}{\nsample}$, $t_0 = \lceil 4\kappa \left( \log 4\kappa +1\right )\rceil$, and $\kappa = 1.1 \frac{\overline{B}}{\underline{B}}$ where 
		$\underline{B}:\,= \inf_{|x|\leq  \alpha'\fronorm{\truX}}\psi''(x)$ and $\overline{B} :\,= \sup_{|x|\leq \alpha'\fronorm{\truX}} \psi''(x)$. 
		\end{lem}
	
	Combined the above lemma and Theorem \ref{thm: mainthm2}, Corollary \ref{cor: distancepairwisesampling} is immediate. 
	Let $r=\trurank$, $\beta = \frac{33}{16}\overline{B}$,  $\eta_0 = \frac{1}{4\kappa}$, and $t_0 =\lceil 4\condn( \log 4\condn +1)\rceil$ to be the input of AVPG.
	
	\begin{cor}[Distance to $\truX$]\label{cor: distancepairwisesampling}
	Instate the assumptions and notation in Lemma \ref{lem: rscrsmSmallGradientPairwiseSampling}, and suppose AVPG 
uses the input described above.
		Define $\tau_\star= \arg\min_{1\leq \tau \leq t+1} \loss(X_t)$. Then  there are 
		some universal constatn $c,c',C$ such that for any  $\nsample <d^2\log \dm$ and $\trurank \nsample \geq C \dm \log \dm $,
		with probability at least $1-c'd^{-2}$ 
		\[
		\fronorm{X_{\tau_\star} - \truX}^2 \leq \frac{c}{\underline{B}}\max\left\{(1-\frac{1}{4\kappa})^{t}(4\kappa)^{s}h_0,{\condn} \log\kappa (\condn^2+\alpha '\sqrt{\overline{B}}\kappa+(\alpha')^2\overline{B})\frac{\trurank\dm\log \dm}{\nsample}\right\}.
		\] 
		Here $s$ is the largest integer so that $st_0\leq t$. 
	\end{cor}
\paragraph{Interpreting Corollary \ref{cor: distancepairwisesampling}}	Same as the case of one-bit matrix completion, for $\alpha'= \bigO(1)$ and $\fronorm{\truX}=\bigO(1)$, the bound reduces to $\bigO(\frac{\trurank \dm\log \dm}{\nsample})$ 
	for $t\rightarrow \infty$ and matches the bound of 
	convex relaxation in \cite{lu2015individualized}.

	\section{Discussion}\label{sec: conclusion}
	
	In this paper, we identify the regularity
	projection oracle as the key component of solving many interesting problems 
	under RGLM. We develop efficient algorithm that converges linearly and globally. Furthermore, we show state-of-art statistical recovery bounds in concrete RGLM problems. Here we lay out a few interesting future directions. 
	\newcontent{
	\begin{itemize}
	\item \textbf{Models beyond RGLM.} We expect that our algorithm and theoretical framework are broadly applicable to other low-rank problems with non-quadratic loss, such as matrix completion 
	with general exponential family response \cite{lafond2015low}, and multinomial sampling scheme as those in \cite{kallus2020dynamic,oh2015collaboratively}.
		\item \textbf{The choice of $t_0$.} Even though our theory requires $t_0$ to be finite, we found that setting $t_0$ to be infinity still works in our experiments. Empirically, this is due to the fact that the iterate becomes very low rank even though the averaging step is performed in every iteration. Also, empirically, we found that setting $t_0=1$, i.e., AVPG is reduced to PG, does not affect the performance of the algorithm, even though our theory requires $t_0 = \bigO(\condn \log \condn)$. It is interesting to theoretically explain why PG is still effective in the RGLM setting. 
		\item \textbf{Overcoming stationary point by larger stepsize.} We found that in the  experiment performed in Figure \ref{fig:wmclargermatrixplot}, even though IHT stays at the nonoptimal stationary point while AVPG is able to get rid of it. Empirically, IHT with a larger stepsize is actually able to escape the fixed point $M_0$. 
	\end{itemize}
}

	\section*{Acknowledgment}
	L. Ding and Y. Chen were partially supported by the National Science Foundation grants CCF-1657420 and CCF-1704828 and the CAREER Award CCF-2047910.

\bibliographystyle{alpha}
\bibliography{references}
\appendix

\section{Proofs and Lemmas for Section \ref{sec: algguarantees}}\label{sec: proofLemmasForalgguarantees}
\subsection{Lemmas for projection oracle \eqref{eq: projectionOracle}}
We denote the Shatten $p$ norm ball with radius $\xi$ as $\mathbb{B}_{S_p}(\xi)$.
\begin{lem}  \label{lem: projection} 
	Let $(u^\star_i,v^\star_i,\sigma_i) \in \real^{\dm_1}\times \real^{\dm_2}\times \real, i=1,\dots,r$ be the  top $r$ left and right singular vectors, and 
	singular  values of $X$. The solution $V^\star$ to the problem  $\min_{\rank(V)\leq r,\,V\in \mathbb{B}_{S_p}(\xi)} \fronorm{X -V}$ is of the following form: 
	$V^\star = \sum_{i=1}^{r}  a_i^\star  u^\star_i(v^\star_i)^\top $ where $a^\star _i\geq 0 $. Here the numbers $a_i^\star ,i=1,\dots,i=r$ 
	are the solution to $\min_{a\in \real^{r},\|a\|_p\leq \xi} \twonorm{a-\sigma}$ where $\sigma = (\sigma_1,\dots,\sigma_r)$.
\end{lem} 
\begin{proof}
	We first note that the solution to the problem $\min_{\rank(V)\leq r,\,V\in \mathbb{B}_{S_p}(\xi)} \fronorm{X -V}$ is of the form 
	$V^\star = \sum_{i=1}^k a_i u_i^\star (v_i^\star)^\top$ by using \cite[Lemma 3.1 and its proof]{allen2017linear}. This means we only need to choose
	$a_i,i=1,\dots, r$. It is then immediate $a=(a_1,\dots,a_r)$ should be the solution to  $\min_{a\in \real^{r},\|a\|_p\leq \xi} \twonorm{a-\sigma}$ and 
	our proof is complete. 
\end{proof}

\subsection{Proof of Theorem~\ref{thm: mainthm2}}
\label{sec:proof_mainthm2}

\begin{proof} 
	We fix $t$ and do a one-step analysis. Consider the following two inequalities for the pair $(X_t,\truX)$:
	\begin{align}
	\frac{\rsc}{2}\fronorm{X_t-\truX}^2          & \geq 2\rscerror, \label{eq: largedistanceRSC}\\ 
	\frac{\rsm\eta^2}{2} \fronorm{X_t-\truX}^2   & \geq \rsmerror, \label{eq: largedistanceRSM}
	\end{align}
	where $\eta =1$ or $\frac{1}{4\kappa}$ depending on whether $t$ is a multiple of $t_0$. Suppose that inequality~\eqref{eq: largedistanceRSC} does not hold, 
	i.e., $\fronorm{X_t-\truX}\leq \frac{4}{\rsc}\rscerror$. In this case,  using 
	the $(\rsmerror,rt_0,\alpha,\conset)$-RSM property, we find that 
	\begin{equation}
	\begin{aligned}\label{eq: rscConditionFailConsequence} 
	h_t=\loss_{\nsample}(X_t)-\loss_{\nsample}(\truX) \leq  & \inprod{\nabla \loss_{\nsample}(\truX)}{X_t-\truX} + 2\condn\rscerror +\rsmerror \\ 
	\overset{(a)}{\leq}   & \nablasmall\sqrt{\frac{8t_0r\rscerror}{\rsc}} +2\condn\rscerror +\rsmerror, 
	\end{aligned}
	\end{equation}
	where in step $(a)$ we use H\"older's inequality and the fact that $X_t,\truX$ has rank no more than $t_0r$. Therefore,  we have the desired bound
	\eqref{eq: thm2eq1}. By a similar argument, we can show that if inequality~\eqref{eq: largedistanceRSM} does not hold, then we again have the desired bound~\eqref{eq: thm2eq1}.
	
	We henceforth assume that both inequalities \eqref{eq: largedistanceRSC} and \eqref{eq: largedistanceRSM} hold. Using the approximate RSM property
	and the update rule of $X_{t+1}$, we find that 
	\begin{equation}
	\begin{aligned}\label{eq: MainStepofThm2}
	\loss(X_{t+1}) 
	& \leq \loss(X_t) + \eta \inprod{\nabla \loss_{\nsample}(X_t)}{V_t-X_t} + \frac{\beta \eta^2}{2}\fronorm{V_t-X_t}^2+ \rsmerror \\ 
	& \overset{(a)}{\leq }  \loss(X_t) + \eta \inprod{\nabla \loss_{\nsample}(X_t)}{\truX-X_t} + \frac{\beta \eta^2}{2}\fronorm{\truX-X_t}^2 + \rsmerror\\
	& \overset{(b)}{\leq}  \loss_{\nsample}(X_t) -\eta\left (\loss_{\nsample}(X_t)-\loss_{\nsample}(\truX) \right)+\eta \left(\beta \eta -\frac{\alpha}{4}\right)\fronorm{\truX-X_t}^2 \\
	& \overset{(c)}{\leq} \loss_{\nsample}(X_t) -\eta h_t +\frac{4\eta }{\alpha} \left({\beta \eta} -\frac{\alpha}{4}\right)  \max\left \{ h_t,\frac{4}{\alpha}( 
	rt_0+\trurank)\nablasmall ^2 \right\}.
	\end{aligned}
	\end{equation}
	Here in step $(a)$, we use the optimality of $V_t$; in step $(b)$, we use the approximate RSC and the inequalities \eqref{eq: largedistanceRSC} and \eqref{eq: largedistanceRSM}; in the last step $ (c) $ we use Lemma \ref{lem: smallfronormX-truX} and \eqref{eq: largedistanceRSC}. Now, we subtract $\loss_{\nsample}(\truX)$  from both sides of \eqref{eq: MainStepofThm2}. Doing so and using  the choice of $\eta$ for $t$ is a multiple of $t_0$ and the cast $t$ is not, we obtain the inequality
	\begin{equation}\label{eq: InductionResult}
	h_{t+1} \leq \begin{cases} 
	(4\kappa-1) \max\left \{ h_t,\frac{4}{\alpha}( 
	r\lceil \kappa t_0+\trurank)\nablasmall ^2 \right\}, &  \exists k\in \integer: \,t=kt_0, \\ 
	(1-\frac{1}{4\kappa})h_t, & \text{otherwise.}
	\end{cases} 
	\end{equation}
	Applying the inequality~\eqref{eq: InductionResult} inductively proves the desired bound~\eqref{eq: thm2eq1} on the objective value. 
	
	Finally, combining the approximate RSC property with the bound on $h_t$ in ~\eqref{eq: thm2eq1} we just proved, we immediately obtain the desired distance bound on $\fronorm{\Delta_t}$ in \eqref{eq: thm2eq1}. 
\end{proof} 

\begin{lem}\label{lem: smallfronormX-truX}
    Given a rank $r$ matrix $X\in \real^{\d_1 \times \d_2}$ with $r\geq \trurank$, and suppose that $\loss$ satisfies 
    $(r,\rsc)$-RSC. Let $\opnorm{\nabla \loss(\truX)} = \nablasmall$. Then we have 
    \[
    \fronorm{X-\truX}^2 \leq\frac{4}{\alpha}\max\left \{ \loss_{\nsample}(X)-\loss_{\nsample}(\truX) ,\frac{4}{\alpha}(r+\trurank)\nablasmall ^2 \right\}.
    \]
\end{lem}
\begin{proof}
	First if $2\abs{\inprod{\nabla \loss_{\nsample}(\truX)}{X-\truX}} \geq \frac{\alpha}{2}\fronorm{X-\truX}^2$, using H\"older's inequality in the following 
	step $(a)$ and $ \nucnorm{X-\truX}\leq \sqrt{r+\trurank}\fronorm{X-\truX}$ in step $(b)$, we have 
	we find that 
	\begin{equation}
	\begin{aligned}\label{eq: lemmasmallfronormX-truXimportant1}
     \frac{\alpha}{2}\fronorm{X-\truX}^2 \leq 2\inprod{\nabla \loss_{\nsample}(\truX)}{X-\truX} & \overset{(a)}{\leq} 2 \opnorm{\nabla \loss (\truX)}\nucnorm{X-\truX} \\ 
      & \overset{(b)}{\leq} 2\sqrt{r+\trurank}\nablasmall \fronorm{X-\truX} \\ 
    \overset{(c)}{\implies} \fronorm{X-\truX} &\leq \frac{4}{\rsc}\sqrt{r+\trurank}\nablasmall.
	\end{aligned}
	\end{equation}
	The step $(c)$ is due to canceling the term $\fronorm{X-\truX}$ from both sides of the inequality. 
	
	Otherwise, we should have $2\abs{\inprod{\nabla \loss_{\nsample}\truX)}{X-\truX} }\leq \frac{\alpha}{2}\fronorm{X-\truX}^2$. 
	Using the $(r,\rsc)$-RSC of $\loss$ in the following step $(a)$, and  $2\abs{\inprod{\nabla \loss_{\nsample}(\truX)}{X-\truX} }\leq \frac{\alpha}{2}\fronorm{X-\truX}^2$
	in step $(b)$, we have 
	\begin{equation}
	\begin{aligned} \label{eq: lemmasmallfronormX-truXimportant2}
	\loss_{\nsample}(X) & \overset{(a)}{\geq} \loss_{\nsample}(\truX) +\inprod{\nabla \loss_{\nsample}(\truX)}{X-\truX}  + \frac{\alpha}{2}\fronorm{X-\truX}^2 \\ 
	& \overset{(b)}{\geq} \loss_{\nsample}(\truX) + \frac{\alpha}{4}\fronorm{X-\truX}^2. 
	\end{aligned} 
	\end{equation}
	By combining  the inequalities \eqref{eq: lemmasmallfronormX-truXimportant1} and \eqref{eq: lemmasmallfronormX-truXimportant2}, we achieve the desired inequality
		\[
		\fronorm{X-\truX}^2 \leq\frac{4}{\alpha}\max\left \{ \loss_{\nsample}(X)-\loss_{\nsample}(\truX) ,\frac{4}{\alpha}(r+\trurank)\nablasmall ^2 \right\}.  
		\]
\end{proof}
\section{Lemmas and proofs for Section \ref{section: Application Examples}}
%

To establish the RSC/RSM and its approximate version for the loss of RGLM, let us first introduce the 
linear operators relating to $A_i,i=1,\dots \nsample$. For any $\mathcal{S}\subset [\nsample]$, we definethe linear operator
$\Amap_{\mathcal{S}}: \real^{\dm_1\times \dm_2} \rightarrow \real^{|\mathcal{S}|}$ 
with 
\[
[\Amap_{\mathcal{S}}(X)]_i = \inprod{\Asense_i}{X}\quad \text{for $i\in \mathcal{S}$}.
\]  
In particular, if $\mathcal{S} = [\nsample]$, we denote $\Amap =\Amap_{[\nsample]}$. 
The corresponding quadratic function of $\Amap_{\mathcal{S}}$ is defined as 
\[
\loss_{\Amap_\mathcal{S}} (X):\,= \frac{1}{2n}\twonorm{\Amap_{\mathcal{S}}(X)}^2.
\]

We shall first show that RSC/RSM, its approximate version of $\loss$ holds whenever
certain deterministic conditions of the map $\Amap_{\mathcal{S}}$. We then verify these two conditions 
holds, as well as small gradient norm condition $\opnorm{\nabla \loss(\truX)}\leq \epsilon_{\nabla}$, with high probability for random $\Amap$ 
so that Theorem \ref{thm: mainthm2}
can be applied.  

\subsection{Deterministic condition of $\Amap_{\mathcal{S}}$ for (approximate) RSC/RSM} 

We require the linear map $\Amap$ to satisfy one of the following properties: 
\begin{itemize}
	\item The function $\loss_{\Amap_{\mathcal{S}}}$ satisfies $(r,\beta,\conset)$-RSC and $(r,\rsc,\conset)$-RSM for each $|\mathcal{S}|\geq  (1-c_0)\nsample$ for some universal $c_0>0$.
	\item  The function $\loss_{\Amap}$ satisfies approximate $(\rscerror,r,\alpha,\conset)$-RSC and $(\rsmerror,r,\beta,\conset)$-RSM.
\end{itemize}

Since the Hessian of $\loss_{\Amap_\mathcal{S}}$ is actually constant, the quadratic form is simply $\nabla ^2 \loss_{\Amap_\mathcal{S}}(Z)[X][X] = 2\loss_{\Amap_\mathcal{S}}(X)$ for 
any $Z\in \real^{\dm_1\times \dm_2}$. The above properties henceforth are easier to establish using techniques in high dimensional probabilities. Specifically, 
we might consider iid Gaussian sensing matrix for the first case, and entrywise type sampling scheme in matrix completion or aggregate individual 
ranking for the second case. 

We shall show that the RSC and RSM of $\loss_{\Amap_{\mathcal{S}}}$ implies the RSC and RSM of $\loss$ with different parameters. Similarly, we show that 
approximate version of RSC and RSM will implies the same properties of $\loss$ with different parameters. However, due to the nonlinearity of 
$\psi'$, we need to restraint our attention to certain bounded set instead of the full space $\real^{\dm_1\times \dm_2}$, and impose certain boundedness assumption on 
$\psi''$.
\begin{lem}\label{lem: gassuainnonlinear}
	Suppose the function $\loss_{\Amap_{\mathcal{S}}}$ satisfies $(r,\rsc,\conset)$-RSC and $(r,\rsm,\conset)$-RSM for any $|\mathcal{S}|\geq (1-c_0)\nsample$ for some universal $c_0>0$
	and $0\in \conset$
	Then the loss $\loss$ satisfies $(\underline{B}\alpha,r, \mathbb{B}_{\fronorm{\cdot}}(\frotruX)\cap \conset)$ RSC and $(r,\overline{B}\beta, \mathbb{B}_{\fronorm{\cdot}}(\xi_0)\cap \conset)$ RSM 
    where $\overline{B} = \infnorm{\psi''}:\,= \sup_{x\in \real}|\psi''(x)|$, and $\underline{B} = \inf_{|x|\leq \sqrt{\frac{2.2\beta}{c_0}} \xi_0} \psi''(x)$ for an $\Amap$ independent $\xi_0>0$.
\end{lem}
\begin{proof}
Given any $X,Y\in \mathbb{B}_{\fronorm{\cdot}}(\frotruX)\cap\conset$ with their ranks  
	not exceeding $r$ and Frobenius norms not exceeding $\xi_0$, define $\Delta :\,= Y-X$. The Taylor expansion of $\loss$ gives 
	\begin{equation}
	\begin{aligned}
	\loss(Y)-\loss_{\nsample}(X) -\inprod{\nabla \loss_{\nsample}(X)}{Y-X} 
	& = \frac{1}{2\nsample} \sum_{i=1}^{\nsample} \psi''(\inprod{X+t_i\Delta}{\Asense_i}) \inprod{\Delta}{\Asense_i}^2 \\
	& \overset{(a)}{\leq} B \frac{1}{2\nsample}\twonorm{\Amap(\Delta)}^2\\
	& \overset{(b)}{\leq} \frac{B\beta}{2} \fronorm{\Delta}^2,                                                                
	\end{aligned}
	\end{equation}
	where $t_i \in [0,1]$. Here we use the assumption $B\geq \infnorm{\psi''}$ in step $(a)$, and the $(r,\beta,\conset)$ RSM of $\loss_{\Amap}$ in step $(b)$. 
	
	To prove restricted strong convexity, we claim that for any $\gamma_0\in(0,1)$, there is 
	at most $\frac{1}{2}\gamma_0\in (0,1)$ fraction of the $\{\inprod{X}{A_i}^2 \}_{i=1}^\nsample$ satisfying 
	$\inprod{X}{A_i}^2 \geq \frac{2.2\beta}{\gamma_0} \fronorm{X}^2$. Indeed, otherwise, we will have $\frac{1}{\nsample} \twonorm{\Amap(X)}^2 \geq \frac{\gamma_0}{2} \frac{2.2\rsm}{
	\gamma_0}\fronorm{X}^2> \beta \fronorm{X}^2$, a contradiction to $(r,\rsm,\conset)$ RSM of $\loss_{\Amap}$ (This is where $0\in \conset$ is used). 
   Similarly, we know that at most $\frac{\gamma_0}{2}$ fraction of the $\{\inprod{Y}{A_i}^2 \}_{i=1}^\nsample$ satisfying 
$\inprod{Y}{A_i}^2 \geq \frac{2.2\beta}{\gamma_0} \fronorm{Y}^2$. Hence we can find a set $\mathcal{S}\subset[\nsample]$ with cardinality  at least
$n(1-\gamma_0)$ such that $\inprod{X}{A_i}^2\leq \frac{2.2\beta}{\gamma_0}\fronorm{X}^2 \leq \frac{2.2\beta}{\gamma_0} \xi_0^2$ for every $i\in \mathcal{S}$ and the same inequality 
holds for $\inprod{Y}{A_i}^2$. By choosing $\gamma_0=c_0$, we see that $\loss_{\Amap_{\mathcal{S}}}$ also satisfies $\alpha$-RSC by our assumption.
Combining pieces, we have 
\begin{equation}
\begin{aligned}
\loss(Y)-\loss_{\nsample}(X) -\inprod{\nabla \loss_{\nsample}(X)}{Y-X} 
& = \frac{1}{2\nsample} \sum_{i=1}^{\nsample} \psi''(\inprod{X+t_i\Delta}{\Asense_i}) \inprod{\Delta}{\Asense_i}^2 \\
& \overset{(a)}{\geq} \underline{B} \frac{1}{2\nsample}\twonorm{\Amap_{\mathcal{S}}(\Delta)}^2\\
& \overset{(b)}{\leq} \frac{ \underline{B}\alpha}{2} \fronorm{\Delta}^2,                                                                
\end{aligned}
\end{equation}
where $t_i \in [0,1]$, and $\underline{B} = \inf_{|x|\leq \sqrt{\frac{2.2\beta}{c_0}} \xi_0} \psi''(x)$. 
Here we use the construction of $\mathcal{S}\subset [\nsample]$ in step $(a)$, and the $\alpha$ RSC of $\loss_{\Amap_{\mathcal{S}}}$ in step $(b)$. 
\end{proof}

\begin{lem}\label{lem: entrywisepairwiseRSCRSM}
	Suppose the function $\loss_{\Amap}$ satisfies $(\rscerror,r,\rsc,\conset)$-RSC and $(\rsmerror,r,\rsm,\conset)$-RSM. If $X\in \conset$ and 
	$\rank(X)\leq r$ implies that $\abs{\inprod{X}{\Asense_i}}\leq \xi_1$ for some $\Amap$ independent $\xi_1>0$. Then $\loss_{\nsample}$ satisfies 
	 $(\underline{B}_1\rscerror,r,\underline{B}_1\rsc,\conset)$-RSC and $(\overline{B}\rsmerror,r,\overline{B}_2\rsm,\conset)$-RSM, where 
	 $\underline{B_1}:\,= \inf_{|x|\leq \xi_1}\psi''(x)>0$ if $\psi$ is strongly convex in 
	 any bounded domain, and $\overline{B}_2 :\,= \sup_{|x|\leq \xi_1} \psi''(x)$.
\end{lem}
\begin{proof}
	Given any $X,Y\in \conset$ with their rank  
	not exceeding $r$, define $\Delta :\,= Y-X$. The Taylor expansion of $\loss$ gives 
	\begin{equation}
	\begin{aligned}\label{eq: quadraticFormEasyCase}
	\loss(Y)-\loss_{\nsample}(X) -\inprod{\nabla \loss_{\nsample}(X)}{Y-X} 
	& = \frac{1}{2\nsample} \sum_{i=1}^{\nsample} \psi''(\inprod{X+t_i\Delta}{\Asense_i}) \inprod{\Delta}{\Asense_i}^2 \\                                                         
	\end{aligned}
	\end{equation}
	where $t_i \in [0,1]$. Using the assumption that $X,Y\in \mathcal{C}$ implies that $\abs{\inprod{X}{A_i}}\leq \xi_1$ and 
	$\abs{\inprod{Y}{A_i}}\leq \xi_1$ for every $i\in [\nsample]$, we see 
	\[
	\abs{\inprod{X+t_i\Delta}{\Asense_i}} \in [\underline{B_1},\overline{B}_2],
	\]
	where $\underline{B_1}:\,= \inf_{|x|\leq \xi_1}\psi''(x)>0$ as $\psi$ is strongly convex in 
	any bounded domain, and $\overline{B}_2 :\,= \sup_{|x|\leq \xi_1} \psi''(x)$. Hence we 
    can combine this inequality with \eqref{eq: quadraticFormEasyCase} and reach that 
    \begin{equation}
    \begin{aligned}\label{eq: quadraticFormEstimate}
   & \frac{\underline{B}_1}{\nsample} \twonorm{\Amap (\Delta)}^2
   & \leq \loss(Y)-\loss_{\nsample}(X) -\inprod{\nabla \loss_{\nsample}(X)}{Y-X} 
   & \leq  \frac{\overline{B}_1}{\nsample} \twonorm{\Amap (\Delta)}^2.\\ 
    \end{aligned}
    \end{equation}
    Using the approximate RSC and RSM properties of the function $\loss_{\Amap}$, we achieved 
    the approximate RSC and RSM properties of $\loss$.
\end{proof}

\subsection{Random $\Amap_{\mathcal{S}}$ satisfying (approximate) RSC/RSM with high probability}
\subsubsection{Gaussian measurements $A_i$} \label{sec: RSCRSMgaussianSensing}
\begin{lem}\label{lem: gauassianRSCRSMsubset}
	If the measurements $A_i$ have iid standard Gaussian entries,
	then for some universal constant $c,c_0,C>0$, so long as $n\geq Crd$, with probability at least $1-\exp(-nc)$,  
	simultaneously for all  $\mathcal{S}\subset [\nsample]$ 
	with $|\mathcal{S}|\geq (1-c_0)\nsample$,  the loss $\loss_{\Amap_{\mathcal{S}}}$ satisfies 
	$(\frac{31}{32},r,\real^{\dm_1\times \dm_2})$-RSC and 	$(\frac{33}{32},r,\real^{\dm_1\times \dm_2})$-RSM. 
	
\end{lem}
\begin{proof}
A standard result \cite[Theorem 2.3]{candes2011tight} shows that
we have $(r,1-\delta,\real^{\dm_1\times \dm_2})$ RSC and $(r,1+\delta,\real^{\dm_1\times \dm_2})$ RSM of 
$\loss_{\Amap_{\mathcal{S}}}$ with probability $1-\exp(-c|\mathcal{S}|)$ for each $\mathcal{S} \subset [\nsample]$ 
if $|\mathcal{S}|\geq c'r \dm$ for some universal $c,c'>0$. 
Let
$|\mathcal{S}|\geq (1-\epsilon )n$ for some 
$\epsilon$ to be determined, there are at most  $\epsilon \nsample {\nsample \choose (1-\epsilon )\nsample}$ 
many $\mathcal{S}$. The number $ \epsilon \nsample {\nsample \choose (1-\epsilon )\nsample} $ is bounded by 
\begin{equation}
\begin{aligned}
                        \epsilon \nsample {\nsample \choose (1-\epsilon )\nsample} 
                 =   &  \exp(\log (\epsilon) + \log \nsample) {\nsample \choose \epsilon \nsample} \\
 \overset{(a)}{\leq} &  \exp(\log \epsilon + \log \nsample) \left (\frac{e}{\epsilon}\right)^{\epsilon \nsample}  \\
               \leq  &  \exp\left(\log\epsilon + \log \nsample) +\epsilon \nsample -(\epsilon\log \epsilon)\nsample \right)\\
               \leq  &   \exp \left(\log\nsample+(\epsilon-\epsilon \log\epsilon )\nsample\right).
\end{aligned}
\end{equation}
Here in step $(a)$ we use the fact that ${\nsample \choose k }\leq \left(\frac{en}{k}\right)^{k}$. Since 
$-\epsilon \log \epsilon \rightarrow 0$ for $\epsilon \rightarrow 0$, we know there are universal constant $c_1>0$, and $c_0,c_3$ depends only 
on $c,c',c_1$ such that 
for every $n>c_1$, and $\mathcal{S}$ with $|\mathcal{S}|\geq (1-c_0)n$ that 
\[
\exp(-c|\mathcal{S}|)\exp \left(\log\nsample+(\epsilon-\epsilon \log\epsilon )\nsample\right) \leq \exp(-c_3\nsample).
\]
The proof is then complete.
\end{proof} 
\subsubsection{Entrywise sampling $A_i$} 
Recall entrywise sampling is defined as follows: the measurement matrix $A_i$ satisfies that $A_i = \sqrt{\dm_1\dm_2} e_{k(i)} e_{l(i)}^\top $. Here for each $i\in [\nsample]$, the number $k(i)\in [\dm_1]$ is uniformly distributed on $[\dm_1]$ independent of anything else, and  
$l(i)$ is uniformly distributed over $[\dm_2]$ and is independent of anything else. Recall the collection of measurement matrices $A_i$, $i=1,\dots,\nsample$ defines our entrywise sampling operator 
$\Amap: \real^{\dm_1\times \dm_2} \rightarrow \real^{\nsample}$ 
with 
$
[\Amap_{\mathcal{S}}(X)]_i = \inprod{\Asense_i}{X}.
$

We have the following lemma from \cite[Theorem 10.17]{wainwright2019high}. 
\begin{lem}\label{lem: entrywiseSampling}
	For the random entrywise sampling operator $\Amap: \real^{\dm_1\times \dm_2} \rightarrow \real^{\nsample}$, let $d= \max\{\dm_1,\dm_2\}$. There are universal constants  
	$c_1,c_2$ that 
	\[
	\left| \frac{1}{n} \frac{\twonorm{\Amap(X)}^2}{\fronorm{X}^2}-1\right| \leq c_1 \alpha_{\tiny \mbox{sp}}(X)\frac{\nucnorm{X}}{\fronorm{X}}\sqrt{\frac{\dm \log \dm}{n}}
	+ c_2 \alpha^2_{\tiny \mbox{sp}} \frac{\dm \log \dm}{\nsample}
	\]
	for all $X\in \real^{\dm_1\times \dm_2}$ with probability at least $1-2e^{-\frac{1}{2}\dm \log \dm}$. 
\end{lem}

\begin{lem}\label{lem: entrywiseSamplingRSCRSM}
	Under the same setting as Lemma \ref{lem: entrywiseSampling}, we have for  with probability at least $1-2e^{-\frac{1}{2}\dm \log \dm}$ that for 
	any $\delta\in (0,1)$, and all $X$ with rank no more than $r$ and $\infnorm{X}\leq \frac{\alpha}{\sqrt{\dm_1\dm_2}}$ simultaneously
		that 
		\begin{equation} \label{eq:  entrywiseSamplingRSCRSM}
		\left| \frac{1}{n} \twonorm{\Amap(X)}^2-{\fronorm{X}^2}\right| 
   \leq 2\delta \fronorm{X}^2
		+ \frac{1}{\delta}c \alpha^2\frac{r\dm \log \dm}{\nsample}.
		\end{equation}
		for some universal $c>0$.
\end{lem}
\begin{proof}
	Using Lemma \ref{lem: entrywiseSampling}, we found that 
	\[
	\left| \frac{1}{n} \twonorm{\Amap(X)}^2-{\fronorm{X}^2}\right| \leq c_1 \sqrt{\dm_1\times\dm_2} \infnorm{X}{\nucnorm{X}}\sqrt{\frac{\dm \log \dm}{n}}
	+ c_2 \dm_1\dm_2 \infnorm{X} \frac{\dm \log \dm}{\nsample}.
	\]
	Combining with $\infnorm{X}\leq \frac{\alpha}{\sqrt{\dm_1\dm_2}}$ and $\rank(X)\leq r$, we have 
	\begin{equation}\label{eq: intermediateinequalityRSCRSMentrywise}
	\left| \frac{1}{n} \twonorm{\Amap(X)}^2-{\fronorm{X}^2}\right| \leq c_1 \alpha {\fronorm{X}}\sqrt{\frac{r\dm \log \dm}{n}}
	+ c_2 \alpha^2\frac{\dm \log \dm}{\nsample}.
	\end{equation}
	Now if $\fronorm{X}^2\leq \frac{1}{\delta}c_1 \alpha {\fronorm{X}}\sqrt{\frac{r\dm \log \dm}{n}}$, or $\fronorm{X}^2\leq \frac{1}{\delta^2}c_2 \alpha^2\frac{\dm \log \dm}{\nsample}$, then we
	always  have 
	for some universal $c_3$ that 
	\[
	\fronorm{X}\leq\frac{1}{\delta} c_3 \alpha\sqrt{\frac{r\dm \log \dm}{\nsample}}.
	\] 
	Combining with \eqref{eq: intermediateinequalityRSCRSMentrywise}, the lemma is immediate. Otherwise, we shall have 
	\[
	\fronorm{X}\geq \max\left\{\frac{1}{\delta}c_1\alpha \sqrt{\frac{r\dm \log \dm}{n}},\frac{1}{\delta\alpha} \sqrt{c_2\frac{\dm \log \dm}{n}}\right\}.
	\]
	The lemma is again immediate by combining the above inequality with  \eqref{eq: intermediateinequalityRSCRSMentrywise}.
\end{proof}
\subsubsection{Pairwise sampling $A_i$}
We consider the sampling scheme described in \cite{lu2015individualized}. Recall the measurement
matrix $A_i$ satisfies that $A_i = \sqrt{\dm_1\dm_2} e_{k(i)}\left(e_{l(i)} -e_{j(i)}\right)^\top$. Here for each $i\in [\nsample]$, the number $k(i)\in [\dm_1]$ is uniformly distributed on $[\dm_1]$ independent of anything else, and  
$(l(i),j(i))$ is uniformly distributed over $[\dm_2]^2$ and is independent of anything else. We shall establish the following lemma. 
\begin{lem}\label{lem: pairwiseSampling}
	For the random pairwise comparison operator $\Amap: \real^{\dm_1\times \dm_2} \rightarrow \real^{\nsample}$, let $d= \max\{\dm_1,\dm_2\}$. If $\nsample <\dm^2\log \dm$, There are universal constants  
	$c_1,c_2$ that 
	\[
	\left| \frac{1}{n} \frac{\twonorm{\Amap(X)}^2}{2\fronorm{X}^2}-1\right| \leq c_1 \alpha_{\tiny \mbox{sp}}(X)\frac{\nucnorm{X}}{\fronorm{X}}\sqrt{\frac{\dm \log \dm}{n}}
		+ c_2 \alpha^2_{\tiny \mbox{sp}} (X)\frac{\dm \log \dm}{\nsample}
	\]
	for all $X\in \real^{\dm_1\times \dm_2}$ with probability at least $1-2e^{-\frac{1}{2}\dm \log \dm}$. 
\end{lem}
\begin{proof}
Let us first define a few notions to ease our proof presentation. As the inequality is homogeneous in $X$, 
we only need to consider $\fronorm{X}=1$. Define the set 
\[
\mathcal{B}(D,\alpha ) = \left\{  X\in \real^{\dm_1\times \dm_2} \mid \fronorm{X}=1,\quad \infnorm{X}\leq \frac{\alpha}{\sqrt{\dm_1 \dm_2}}, \quad \text{and}\quad \nucnorm{X}\leq D  \right\}.
\]
Let $F_{X}(A) = \inprod{X}{A_i}^2$.
$M(D) = \sup_{X \in \mathcal{B}(D)} \left | \frac{1}{n} \sum_{i=1}^n F_{X}(A) -\Exs[F_{X}(A)]\right |$.
Note that   
\begin{equation}
\begin{aligned} 
\Exs [F_X(A)]  & = \dm_1\dm_2 \Exs\left [\left(X_{k(i)l(i)}-X_{k(i)j(i)}\right)^2\right ] \\
& = \frac{1}{\dm_2}\sum_{1\leq k \leq \dm_1,1\leq j,l\leq \dm_2} (X_{kl}-X_{kj})^2 \\
& = \frac{1}{\dm_2}\left( \sum_{1\leq k \leq \dm_1,1\leq j,l\leq \dm_2} X^2_{kl} +  \sum_{1\leq k \leq \dm_1,1\leq j,l\leq \dm_2} X^2_{kj} - \sum_{1\leq k \leq \dm_1,1\leq j,l\leq \dm_2}2 X_{kj}X_{kl}\right) \\
& \overset{(a)}{=} 2\fronorm{X}^2 -\frac{1}{\dm_1} \sum_{1\leq k\leq d_1 ,1\leq j\leq \dm_2} X_{kj}\underbrace{\sum_{1\leq l\leq \dm_2} X_{kl}}_{=0}. \\
& = 2\fronorm{X}^2.
\end{aligned}
\end{equation} 
Here in step $(a)$, we use the fact that $X\in \mathcal{B}(D)$ implies that the row sum of $X$ is zero for each row.

We shall now prove the lemma via the standard argument: concentration around the mean, bounding expectation, and the peeling argument. 
Denote $\|B\|_1= \sum_{1\leq i\leq \dm_1,1\leq j\leq \dm_2}|B_{ij}|$ be the vector $\ell_1$ norm on any matrix $B\in \real^{\dm_1\times \dm_2}$.   

For the concentration around the mean, we first find that 
\begin{equation}
\begin{aligned}\label{eq: pairwisebernsteinboundedConcentration}
|F_{X}(A_i)| \overset{(a)}{\leq} \infnorm{X}^2 \|A_i\|_1^2 \overset{(b)}{\leq} \alpha^2.     
\end{aligned}
\end{equation}
Here we use the H\"older's inequality in step $(a)$ and the $\infnorm{X}\leq \frac{\alpha}{\sqrt{\dm_1\dm_2}}$, and the definition of $A_i$ in step $(b)$.  
For the variance of $F_{X}(A_i)$, we have 
\begin{equation}
\begin{aligned}\label{eq: pairwisebernsteinvarConcentration}
\mbox{var}(F_{X}(A_i)) \leq \Exs [F_X^2(A_i) ]\overset{(a)}{\leq}\alpha ^2   \Exs [F_X(A_i) ] =2\alpha^2.     
\end{aligned}
\end{equation}
Here in step $(a)$, we use \eqref{eq: pairwisebernsteinboundedConcentration}. Combining the 
inequalities \eqref{eq: pairwisebernsteinboundedConcentration} and \eqref{eq: pairwisebernsteinvarConcentration}, 
using the Talagrand concentration for empirical process in Lemma \ref{lem: functionalBernstein} with $\epsilon=1$ and $t= \frac{\dm \log\dm}{\nsample}$, 
we conclude that 
there are some universal constants $c_1,c_2$ such that 
\begin{equation}\label{eq: pairwisebernsteinConcentration}
\begin{aligned} 
\Prob\left[ M(D,\alpha)\geq 2\Exs M(D,\alpha) + \frac{c_1}{8} \alpha \sqrt{\frac{\dm \log \dm}{n}} + \frac{c_2}{4}\alpha^2 \frac{\dm \log\dm}{\nsample}  \right]  
\leq \exp(-\dm \log \dm). 
\end{aligned} 
\end{equation}

For bounding the expectation $\Exs M(D,\alpha)$, by following the proof in \cite[Lemma 3]{lu2015individualized} with minor modification (this 
is where the condition $n<\dm^2\log\dm$ used), we find that 
\begin{align}\label{eq: pairwisebernsteinBoundingExpectation}
\Exs M(D,\alpha)                               & \leq \frac{c_1}{16}\alpha D\sqrt{\frac{\dm \log\dm}{n}}. 
\end{align} 
for some appropriate chosen universal constant $c_1$. 

Finally, we shall use a peeling argument to prove the bound for all $\alpha $ and $D$. Note our bounds \eqref{eq: pairwisebernsteinvarConcentration},
and \eqref{eq: pairwisebernsteinConcentration} match exactly the ones in \cite[Proof of Theorem 10.17]{wainwright2019high}, using the 
step \cite[Extension via peeling]{wainwright2019high} there, we conclude our lemma.  

\end{proof}

Using Lemma \ref{lem: pairwiseSampling} and the proof of Lemma \ref{lem: entrywiseSamplingRSCRSM}, the following 
lemma for approximate RSC/RSM is immediate. 
\begin{lem}\label{lem: pairwiseSamplingRSCRSM}
	Under the same setting as Lemma \ref{lem: pairwiseSampling}, we have for  with probability at least $1-2e^{-\frac{1}{2}\dm \log \dm}$ that for 
	any $\delta\in (0,1)$, and all $X$ with rank no more than $r$ and $\infnorm{X}\leq \frac{\alpha}{\sqrt{\dm_1\dm_2}}$ simultaneously
	that 
	\begin{equation} \label{eq:  pairwiseSamplingRSCRSM}
	\left| \frac{1}{n} \twonorm{\Amap(X)}^2-2{\fronorm{X}^2}\right| 
	\leq 2\delta \fronorm{X}^2
	+ \frac{1}{\delta}c \alpha^2\frac{r\dm \log \dm}{\nsample}.
	\end{equation}
	for some universal $c>0$.
\end{lem}

\begin{lem}[Talagrand concentration for empirical process]\cite[Theorem 3.27, and Equation (3.86)]{wainwright2019high}\label{lem: functionalBernstein}
	Consider a countable class of functions $\mathcal{F}:\mathcal{X}\rightarrow \real$ uniformly bounded by $b$, where 
	$\mathcal{X}\subset \real^\dm$ for some $\dm$. For a series of i.i.d. random variable 
	$X_i$ follows probability distribution $\Prob_X$ supported on $\mathcal{X}$. Define $\sigma^2 = \sup_{f\in \mathcal{F}} \Exs f(X)$. Then 
	for any $\epsilon,t>0$, tje ramdp, variable $Z = \sup_{f\in\mathcal{F}}\frac{1}{\nsample}\sum_{i=1}^\nsample f(X_i)$ satisfies the upper tail bound
	\[
	\Prob[Z\geq (1+\epsilon)\Exs Z +c_0\sigma \sqrt{t} +(c_1+c_0^2/\epsilon) bt ] \leq e^{-nt},
	\]   
	for some universal constant $c_0>0$.
\end{lem}

\subsection{Proof of approximate RSC/RSM for Gaussian measurements, entrywise sampling, and pairwise sampling}

\subsubsection{Proof of approximate RSC/RSM for Gaussian measurements of Lemma \ref{lem: rscrsmsmallgradientgaussianmearsurement}}\label{sec: proofOfrscrsmsmallgradientgaussianmearsurement}
The approximate RSC and RSM condition  listed in Lemma \ref{lem: rscrsmsmallgradientgaussianmearsurement} is 
immediate by combining Lemma \ref{lem: gauassianRSCRSMsubset} and \ref{lem: gassuainnonlinear}. 

\subsubsection{Proof of approximate RSC/RSM for entrywise measurements of Lemma~\ref{lem: rscrsmsmallgradientEntrywiseSampling}}\label{sec: proofOfrscrsmsmallgradientEntrywisemearsurement}
The approximate RSC and RSM condition listed in Lemma \ref{lem: rscrsmsmallgradientEntrywiseSampling} is 
immediate by combining Lemma \ref{lem: entrywiseSamplingRSCRSM} and \ref{lem: entrywisepairwiseRSCRSM}. 

\subsubsection{Proof of approximate RSC/RSM for entrywise measurements of Lemma~\ref{lem: rscrsmSmallGradientPairwiseSampling}}\label{sec: proofOfrscrsmsmallgradientPairwisemearsurement}
The approximate RSC and RSM condition listed in Lemma \ref{lem: rscrsmSmallGradientPairwiseSampling} is 
immediate a simple consequence of combining Lemma \ref{lem: entrywisepairwiseRSCRSM} and \ref{lem: pairwiseSamplingRSCRSM}.
\subsection{Small gradient norm $\opnorm{\nabla \loss(\truX)}$}
\subsubsection{Gaussian measurements $A_i$}
Our first lemma draws the connection between the gradient $\nabla \loss(\truX)$ and the map $\Amap$. 
\begin{lem}\label{eq: subgaussianNoise}
	For the exponential family noise model in \eqref{eq: glm}, we have 
	\[
	\nabla \loss (\truX) = \frac{1}{\nsample} \sum_{i=1}^{\nsample} \left(\psi'(\inprod{\truX,A_i}) -y_i\right) \Asense_i.
	\]
	If $\bar{B}:\, = \infnorm{\Psi''}<\infty$, then each $w_i :\,=\psi'(\inprod{\truX,A_i}) -y_i$ is subgaussian conditional on $A_i$ with $ \Exs(\exp(tw_i) \mid A_i)\leq \exp(\frac{t^2\overline{B}}{2c(\sigma)})$.
\end{lem}
\begin{proof}
The formula for $\nabla \loss(\truX)$ is immediate given the definition of $\loss$. To show $w_i$ is subgaussian, denote the shorthand 
that $\theta_i = \inprod{A_i}{\truX}$. Then 
\begin{equation}
\begin{aligned}
\log \Exs(\exp(tw_i) \mid A_i)  & = t\psi'(\theta_i) +\frac{1}{c(\sigma)}\left( \psi(\theta_i -tc(\sigma)) - \psi(\theta_i) \right) \\ 
                                & \leq \frac{1}{2c(\sigma)}t^2c^2(\sigma)\psi''(\theta_i -\tilde{t} c(\sigma)) \\ 
                                & \leq \frac{1}{2c(\sigma)}t^2c^2(\sigma)\overline{B}. 
\end{aligned} 
\end{equation}
\end{proof}
\begin{lem}\label{lem: gaussiangradienNorm}
	Suppose the sensing scheme is Gaussian where each $A_i$ has i.i.d. standard Gaussian entries. Let   
	$d = \max\{\dm_1,\dm_2\}$. Then the following bound holds 
	\[
	\Prob\left( \opnorm{\nabla \loss(\truX)}\leq \sqrt{c(\sigma)\overline{B}} \sqrt{\frac{\dm}{\nsample}}\right)\leq  1-\exp(-c\dm),
	\]
	where $c$ is some universal constant, and $\bar{B}:\, = \infnorm{\Psi''}$. 
\end{lem}
\begin{proof}
	Let $Q = \nabla \loss_{\nsample}(\truX)$. Consider $[u^1,\dots, u^M]$ and $[v^1,\dots,v^N]$ be $1/4$-covers in Euclidean norm of the spheres 
	$\mathbb{S}^{\dm_1-1}$ and $\mathbb{S}^{\dm_2-1}$, respectively. By lemma \cite[Lemma 5.7]{wainwright2019high}, we know 
	we can make $M\leq 9^{\dm_1}$ and $N\leq 9^\dm_2$. Standard covering argument (see for example \cite[page 324]{wainwright2019high}) shows that 
	\begin{equation}\label{eq: qlessunionuv}
	\opnorm{Q}\leq 2\max_{1\leq j\leq M,1\leq l\leq N}Z^{j,l},\quad \text{where}\; Z^{j,l}=\inprod{u^j,Qv^l}.
	\end{equation}
We can decompose $Z^{j,l}$ as $Z^{j,l} = \frac{1}{n}\sum_{i=1}^n w_i Y^{j,l}_i$ where $w_i =\psi'(\inprod{\truX,A_i}) -y_i$, and $Y ^{j,l}_i = \inprod{u^j}{A_iv^l}$. 
 Since $w_i$ and $Y^{j,l}$ are subgaussian with parameter $\sqrt{c(\sigma)\overline{B}}$ and $1$ respectively \cite[Definition 2.5.6]{vershynin2018high}, 
 we know that $w_i Y^{j,l}_i$ is subexponential with parameter $K:=\sqrt{c(\sigma)\overline{B}}$ \cite[Definition 2.7.5, Lemma 2.7.7]{vershynin2018high}. Using the Bernstein's 
 inequality \cite[Corollary 2.8.3]{vershynin2018high}. we have that 
 \begin{equation}
 \Prob\left\{ \left | \frac{1}{n}\sum_{i=1}^n w_i Y^{j,l}_i \right|\geq t\right\}\leq 2\exp \left(-nc\min\left( \frac{t^2}{K^2},\frac{t}{K}\right)\right).
 \end{equation}
Taking $t= CK\sqrt{\frac{d}{n}}$ for some universal constant $C>0$, we find that with probability at least 
\begin{equation}\label{eq: Zjlbound}
 \Prob\left(|Z^{j,l}| = \left |\frac{1}{n}\sum_{i=1}^n w_i Y^{j,l}_i \right | \leq CK\sqrt{\frac{d}{n}}\right) \leq 1- 2\exp(-9\dm). 
\end{equation}
A union bound on all $u^j$ and $v^l$ shows that previous inequality holds with probability at least $1-2\exp(-c\dm)$ 
for some universal $c>0$ simultaneously for all $u^j,v^l$, $j=1,\dots,M$, and $l=1,\dots,N$. Hence, combining inequalities 
\eqref{eq: Zjlbound} and \eqref{eq: qlessunionuv},we find that with probability at least $1-\exp(-c\dm)$, we have 
\[
\opnorm{\nabla \loss(\truX)}\leq C\sqrt{c(\sigma)\overline{B}} \sqrt{\frac{\dm}{\nsample}}.
\]
\end{proof} 

Let us consider the noise family is Gaussian, $\psi(\theta)= \frac{1}{2}\theta^2$ and $c(\sigma)=\sigma^22$, or 
is Bernoulli, $\psi(\theta) = \log(1+e^\theta)$ and $c(\sigma)=1$:  
\begin{itemize}
	\item Gaussian noise: $\overline{B}=1$, and w.h.p. 
	\begin{equation}\label{eq: gaussiansensinggaussiannoisegradientnormatgroundtruth}
	\opnorm{\nabla \loss(\truX)}\leq c\sigma \sqrt{\frac{\dm}{\nsample}}.
	\end{equation}
	\item Bernoulli noise: $\overline{B}\leq 2$, and w.h.p. 
	\begin{equation}\label{eq: gaussiansensingBernoullinoisegradientnormatgroundtruth}
	\opnorm{\nabla \loss(\truX)}\leq C\sqrt{\frac{\dm}{\nsample}}.
	\end{equation}
\end{itemize} 
\subsubsection{Entrywise and pairwise sampling $A_i$}
Here we assume RGLM is either Bernoulli response, $c(\sigma) =1$, $\psi(\theta) =\log(1+e^\theta)$,
 or Gaussian response, $\psi(\theta)= \frac{1}{2}\theta^2$ and $c(\sigma)=\sigma^2$. We show the following Lemma 
 for Entrywise and pairwise sampling. 
 \begin{lem}\label{lem: smallnormGradientpairEntrywise}
 For Bernoulli response and Gaussian response of RGLM with entrywise sampling or
 Bernoulli response of RGLM with pairwise sampling, 
 there exists universal constant $c>0$ such that with probability 
 at least $1-\dm^{-2}$, there holds 
 $\opnorm{\nabla \loss (\truX)} =\opnorm{\frac{1}{\nsample} \sum_{i=1}^{\nsample} \left(\psi'(\inprod{\truX,A_i}) -y_i\right) \Asense_i} \leq  c\sqrt{c(\sigma)\frac{\dm \log \dm}{n}}$.
 \end{lem}
\begin{proof} 
	We consider the Bernoulli response and Gaussian response seperately.
\begin{itemize}
	\item For entrywise or pairwise sampling scheme with Bernoulli noise, a direct application of Lemma \ref{lem: AW} yields with probability at 
	lest $1-\frac{2}{\dm^2}$: 
	\begin{equation}\label{eq: smallnormgradiententrywisepairwisebernoulli}
	\opnorm{\nabla \loss (\truX)} =\opnorm{\frac{1}{\nsample} \sum_{i=1}^{\nsample} \left(\psi'(\inprod{\truX,A_i}) -y_i\right) \Asense_i} \leq  8\sqrt{\frac{\dm \log \dm}{n}},
	\end{equation}
	where $d=\max(\dm_1,\dm_2)$. 
	\item For entrywise  sampling scheme with Gaussian noise, utilizing \cite[Example 6.18]{wainwright2019high}, we find that with probability at least $1-d^{-2}$,  
		\begin{equation}\label{eq: smallnormgradiententrywisepairwiseGaussian}
	\opnorm{\nabla \loss (\truX)} =\opnorm{\frac{1}{\nsample} \sum_{i=1}^{\nsample} \left(\psi'(\inprod{\truX,A_i}) -y_i\right) \Asense_i} \leq  8\sigma \sqrt{\frac{\dm \log \dm}{n}},
	\end{equation}
	where $d=\max(\dm_1,\dm_2)$.
\end{itemize}
 \end{proof}  
\begin{lem}\cite[Theorem 1.6]{tropp2012user} \label{lem: AW} Let $W_i$ be independent $\dm_1\times \dm_2$ zero-mean random matrices such that $\opnorm{W_i}\leq M$, and define 
	$\sigma_i ^2 :\,= \max\{ \opnorm{\Exs [W_i^\top W_i]},\opnorm{\Exs[W_iW_i^\top]}\}$ as well as $\sigma = \sum_{i=1}^\nsample \sigma_i^2$. We have 
	\[
	\Prob \left[ \opnorm{\sum_{i=1}^nW_i} \geq t \right]\leq (\dm_1+\dm_2)\max\left\{\exp (-\frac{t^2}{4\sigma^2}),\exp \left(-\frac{t}{2M}\right)\right\}.
	\]  
\end{lem}
The following lemma seems to be convenient for Poisson and exponential case. 
\begin{lem}\cite[Proposition 21]{lafond2015low} \cite[Proposition 11]{klopp2014noisy}\cite[Theorem 4]{koltchinskii2011nuclear}
Consider a finite sequence of independent random matrices $(Z_i)_{1\leq i\leq \nsample}\in \real^{\dm_1\times \dm_2}$ satisfying $\Exs[Z_i]=0$. For some $U>0$, assume 
\[
\inf\{\delta >0\mid \Exs [\exp(\opnorm{Z}/\delta )] \leq e\}\leq U \quad \text{for}\quad i=1,\dots,\nsample,
\]
and define $\sigma_Z$ as $\sigma_Z^2 = \max\{ \opnorm{\frac{1}{n}\Exs [Z_i^\top Z_i]},\opnorm{\frac{1}{n}\Exs[Z_iZ_i^\top]}\}$. Then for any $t>0$ with probability at least 
$1-e^{-t}$, 
\[
\opnorm{\frac{1}{n} \sum_{i=1}^\nsample Z_i } \leq c_U \max \left\{ \sigma_Z \sqrt{\frac{t+\log \dm}{\nsample}} , U\log \left( \frac{U}{\sigma_Z}\right)\frac{t+\log \dm}{\nsample} \right\},
\] with $d= \max\{\dm_1,\dm_2\}$ and $c_U$ a constant which depends only on $U$. 
\end{lem}

\section{Additional numerics} \label{sec: numerics}
Here we described the experiments for one-bit matrix completion.
\paragraph{Problem simulation setup} We simulate the ground truth via $\truX=\frac{M_1M_2}{0.3\infnorm{M_1M_2}}$ where each entry of $M_1\in \real^{\dm_1\times r}$ and $M_2\in  \real^{\dm_2\times r}$ is drawn from uniform distribution on $[-0.5,0.5]$ independently.
Instead of the entrywise sampling scheme described in Section \ref{sec: matrixCompletionexponentialfamily}, we use Bernoulli sampling. Given a number $p\in [0,1]$, for each index $(i,j)\in [\dm_1]\times [\dm_2]$,
we observe $(y_{ij},w_{ij})\in \{0,1\}^2$ where $y_{ij} = z_{ij}w_{ij}$ where $z_{ij} \sim \text{Bernoulli}(\frac{1}{1+\exp(-\truX_{ij})})$ independent of anything else and $w_{ij} \sim\text{Bernoulli}(p)$ independent 
of anything else. The Bernoulli sampling is mainly a convenience of the implementation and actually the one originally studied by \cite{davenport20141}. 
We should consider the sample size as $n=p\dm_1\dm_2$ here. We set $\dm_1=\dm_2 =100$, $p=0.5$, and $r=1$ in our experiment.

\paragraph{The PG algorithm and heuristic setup} Next, we perform the PG algorithm \eqref{eq: svp} with the regularity oracle $\proj_{r,\mathbb{B}_{\infnorm{\cdot}}(\infnorm{\truX})}$ and the simple $r$-SVD $\proj_{r,\real^{\dm_1\times \dm_2}} $ starting 
at different random initialization $X_0$. More specifically, we simulate $X_{-1}=\frac{M_{-1}M_{-2}}{0.5\infnorm{M_{-1}M_{-2}}}$ where each entry of $M_i$, $i=-1,\,-2$, is drawn from uniform distribution on $[-0.5,0.5]$ independently, and then set $X_0 = \gamma X_{-1}$ with $\gamma = 0,\,1,\,2,\,4$. Note that $\proj_{r,\mathbb{B}_{\infnorm{\cdot}}(\infnorm{\truX})}$ cannot be computed 
efficiently to our best knowledge. Hence we use the heuristic $\proj_{r,a_u,a_v}$, a modified version of alternate projection, described as follows: Choose two positive number $a_u$ and $a_v$. Given input $X$, 
we first compute the $r$-SVD of $X$ as $U\Sigma V^\top$. Then we compute $U^1=U\sqrt{\Sigma}$ and $V^1=V\sqrt{\Sigma}$ where the square root is applied entrywisely. Next, for each 
$i=1,\dots,\dm_1$ we perform the following operation for the $i$-th
row $U_{i\cdot}^1$ of  $U^1$: 
\[
U^{1}_{i\cdot} \leftarrow \begin{cases}
U^{1}_{i\cdot} & \text{if}\; \twonorm{U^{1}_{i\cdot}}\leq a_u, \\
a_u\frac{U^{1}_{i\cdot}}{\twonorm{U^{1}_{i\cdot}}} & \text{otherwise}.\\
\end{cases}
\]  
Similarly, we perform the following operation for each row of $V^1_j$:
\[
V^{1}_{j\cdot} \leftarrow \begin{cases}
V^{1}_{j\cdot} & \text{if}\; \twonorm{V^{1}_{j\cdot}}\leq a_v, \\
a_v\frac{V^{1}_{j\cdot}}{\twonorm{V^{1}_{j\cdot}}} & \text{otherwise}.\\
\end{cases}
\]  
We then set the output of $\proj_{r,a_u,a_v}$ as $U^1V^1 = \proj_{r,a_u,a_v}(X)$.
We use this heuristic $\proj_{r,a_u,a_v}$ to replace  $\proj_{r,\mathbb{B}_{\infnorm{\cdot}}(\infnorm{\truX})}$ in our PG algorithm. The choice of $a_u$ and $a_v$ is chosen according the 
$r\text{-}SVD$ of $\truX = U^\natural \Sigma ^{\natural} (V^\natural)^\top$. We choose $a_u = \max_{1\leq i\leq \dm_1} \twonorm{U^{\natural,1}_{i\cdot}}$ and 
$a_v=\max_{1\leq j\leq \dm_2} \twonorm{V^{\natural,1}_{j\cdot}}$, where $U^{\natural,1}=U^{\natural}\sqrt{\Sigma^\natural}$ and $V^{\natural,1}=V^{\natural}\sqrt{\Sigma^\natural}$.

\section{Discussion on condition number of one-bit matrix completion} \label{sec: furhterdisccussionOnConditionNumber} 
We consider the condition number of the loss $\loss$ in the one-bit matrix completion setting. We first 
show that condition number is unbounded without $\conset$ being a infinity norm ball and hence 
theoretical guarantees of IHT does not apply. 
We next argue that even in certain favorable setting with $\conset$, where convex relaxation \cite{davenport20141} and 
our AVPG succeed, the results for Burer-Monteiro approach 
in \cite{ge2017no,zhu2018global, zhang2018primal} are still not applicable.

Let us consider the population loss 
\begin{equation}\label{eq: ExpectedLossFormulaOneBit}
\bar{\loss} = \Exs\loss = \frac{1}{\dm_1\dm_2}\left( \sum_{1\leq i\leq \dm_1,\,1\leq j\leq \dm_2}\psi(1+\exp(\sqrt{\dm_1\dm_2}X_{ij})) - \frac{\sqrt{\dm_1\dm_2}
X_{ij}}{1+\exp\left(-\truX_{ij}\sqrt{\dm_1\dm_2}\right)}\right).
\end{equation}
The loss has unbounded condition number over all matrices as  $\psi''(\theta) \rightarrow 0$ as $\theta\rightarrow\pm  \infty$ using 
the equation \eqref{eq: ExpectedLossHessianFormulaOneBit}.

Now we argue the results for Burer-Monteiro approach 
in \cite{ge2017no,zhu2018global, zhang2018primal} are not applicable to one-bit matrix completion even if $\conset$ is present. 
First, if the results in \cite{ge2017no,zhu2018global, zhang2018primal} is not applicable to the population loss, one should 
not expect they can be applied to the sample version $\loss$. Next, it is fairly 
obvious from \eqref{eq: ExpectedLossFormulaOneBit}, the expected loss is not a quadratic and hence result in \cite{ge2017no} don't 
apply. 
Let us now explain what we mean by favorable setting. Consider the Hessian for any $X,\Delta \in \real^{\dm_1,\dm_2}$
\begin{equation}\label{eq: ExpectedLossHessianFormulaOneBit}
\Exs (\nabla ^2\loss_{n}(X)[\Delta,\Delta]) =\sum_{1\leq i\leq \dm_1,1\leq j\leq \dm_2} \psi''(\sqrt{\dm_1\dm_2}X_{ij}) \Delta _{ij}^2.
\end{equation}
We would like to have $\psi''(\sqrt{\dm_1\dm_2}X_{ij}) \in [c_1,c_2]$ for some universal positive $c_1$ and $c_2$, 
so that $\Exs (\nabla ^2\loss_{n}(X)[\Delta,\Delta]) \in [c_1\fronorm{\Delta}^2,c_2\fronorm{\Delta}^2$. 
That is, the Hessian behaves like a quadratic up to some constants. Note that the condition number is simply $\frac{c_2}{c_1}$ here. This 
scenario is actually implied from $\fronorm{\truX}\asymp 1$ and 
$\alpha_{\tiny \mbox{sp}}(\truX)\asymp 1$ when the constraint $\conset$ is $\mathbb{B}_{\infnorm{\cdot}}(\frac{\alpha_{\tiny \mbox{sp}}\fronorm{\truX}}{\sqrt{\dm_1\dm_2}})$.
Hence we define the favorable case to be the setting of one-bit matrix completion with $\fronorm{\truX}\asymp 1$,
$\alpha_{\tiny \mbox{sp}}(\truX)\asymp 1$, and the constraint $\conset$ is $\conset=\mathbb{B}_{\infnorm{\cdot}}(\frac{\alpha_{\tiny \mbox{sp}}\fronorm{\truX}}{\sqrt{\dm_1\dm_2}})$.

We recall our interpretation after Corollary \ref{cor: distanceentrywisesampling} that AVPG produces the distance bound $\bigO\left( \frac{(\trurank)^2\dm \log \dm}{\nsample}\right)$ 
under these favorable conditions. Results in \cite{zhu2018global, zhang2018primal} require the condition number very close to $1$ and more concretely $1.5$ in \cite{zhu2018global} and $\frac{18}{17}$ in \cite{zhang2018primal}. 
Also such constant is not an artifact of the proof as shown in \cite[pp. 3-4]{zhu2018global}. Once the condition number $\condn>3$, it is possible to have spurious local minima. 
Does $\bar{\loss}$ satisfies the condition number less than or equal to $1.5$ in the favorable setting? The answer is no in general, even if $\fronorm{\truX}\asymp 1$,
$\alpha_{\tiny \mbox{sp}}(\truX)\asymp 1$. To see this, consider $\truX = \frac{\gamma}{\sqrt{\dm_1\dm_2}}J$ for some $\gamma\geq 1$, $\alpha_{\tiny \mbox{sp}}(\truX)= 1$ and 
$\conset=\mathbb{B}_{\infnorm{\cdot}}(\frac{\gamma}{\sqrt{\dm_1\dm_2}})$, where $J\in \real^{\dm_1\times \dm_2}$ is the all one matrix with rank one. For constant $\gamma$ independent of the dimension, 
the one-bit matrix completion is in the favorable case. And $\psi''(\sqrt{\dm_1\dm_2}X_{ij}) \in [c_1,c_2]$ for some dimension dependent constant $c_1,c_2$
for any $X\in \conset$. However, this number $c_2/c_2$
can be larger than $1.5$ when  $\gamma \geq 10$. More concretely, set $\gamma =10$, and  consider $X_1= 0.1\frac{1}{\sqrt{\dm_1\dm_2}}e_1e_1^\top$ and 
$X_2 = 5\frac{1}{\sqrt{\dm_1\dm_2}}e_1e_1^\top$, and $\Delta =\frac{1}{\sqrt{\dm_1\dm_2}}e_1e_1^\top$. 
For these $X_i,\,i=1,\,2$ and $\Delta$, they all belong to the set $\conset$, and the equation \eqref{eq: ExpectedLossHessianFormulaOneBit} reduces to 
\begin{align}
\Exs (\nabla ^2\loss_{n}(X_1)[\Delta,\Delta]) &\in [0.24\fronorm{\Delta}^2,0.25\fronorm{\Delta}^2],\quad\text{and}\label{eq: ExpectedLossHessianFormulaOneBitSpecialCaseEqX1}\\
\Exs (\nabla ^2\loss_{n}(X_2)[\Delta,\Delta]) &\in [0.006\fronorm{\Delta}^2,0.007\fronorm{\Delta}^2] \label{eq: ExpectedLossHessianFormulaOneBitSpecialCaseEqX2}.
\end{align}
Thus the condition number is at least $\frac{0.24}{0.007}>1.5$. Hence the results in \cite{zhu2018global, zhang2018primal} don't really apply to this favorable case.

%

\end{document}